\documentclass[10pt,twocolumn,letterpaper]{article}
\usepackage[pagenumbers]{cvpr}
\usepackage[T1]{fontenc}
\usepackage{times}
\usepackage{graphicx,amsmath,amssymb,booktabs,multirow}
\usepackage{colortbl,tabularx}
\newcolumntype{L}[1]{>{\hsize=#1\hsize\linewidth=\hsize\raggedright\arraybackslash}X}
\newcolumntype{C}[1]{>{\hsize=#1\hsize\linewidth=\hsize\centering\arraybackslash}X}
\newcommand{\tablehead}[1]{\textbf{\boldmath #1}}
\usepackage{microtype}
\renewcommand{\floatpagefraction}{0.80}

\usepackage{enumitem}
\usepackage{algorithm,algpseudocode}
\usepackage{amsthm}
\definecolor{paperteal}{HTML}{278E85}
\definecolor{paperpurple}{HTML}{8065A5}
\newcommand{\method}{GeoReVis}
\newcommand{\IoU}{\operatorname{IoU}}

\newtheorem{proposition}{Proposition}

\definecolor{paperblue}{rgb}{0.21,0.49,0.74}
\usepackage[pagebackref,breaklinks,colorlinks,allcolors=paperblue]{hyperref}
\title{GeoReVis: Learning Selective Revision and Evidence Consolidation\\for Remote Sensing Segmentation}
\author{Ke An\quad Liang Chen\quad Yupei Wang\thanks{Corresponding author.}\\
Beijing Institute of Technology}
\hypersetup{pdfauthor={Ke An, Liang Chen, Yupei Wang}}

\begin{document}
\addtocontents{toc}{\protect\setcounter{tocdepth}{-1}}
\raggedbottom
\maketitle
\begin{abstract}
Recent methods try to inspect and revise masks to mitigate referential ambiguity and localization errors caused by complex spatial relations in language-guided remote sensing segmentation. However, these solutions often lead to \textbf{Revision Degradation}: revising missing regions may introduce errors into previously correct ones. Even a globally better mask can omit correct regions found in inferior masks. We term this \textbf{Evidence Fragmentation}: correct target evidence is scattered across imperfect masks. All masks generated by the model at this time are suboptimal.
We propose \textbf{GeoReVis}, which unifies revision guided by execution feedback, Candidate Selection, and Cross-Mask Evidence Consolidation.
It learns fine-grained revision from real actions that improve segmentation when executed and learns to select a better mask through quality comparisons between actual candidates, reducing the risk of degradation during revision. The selected whole mask can still discard useful regions. GeoReVis therefore retains all candidates across rounds, partitions their spatial disagreements into composable regions, and jointly reconstructs regions conditioned on the full instruction and context. This makes complementary and fragmentary evidence reusable regardless of its source candidate's overall quality, yielding improved segmentation.
We further build a data pipeline to generate fine-grained multi-round segmentation trajectories and derive supervision for revision actions, candidate selection, and region composition from the model's own outputs, extending training beyond whole-mask IoU supervision. We also propose \textbf{ReVis-Eval}, a benchmark protocol evaluating multi-round revision and cross-round evidence consolidation accuracy. Experiments on ReVis-Eval and 5 benchmarks for remote sensing reasoning and referring segmentation demonstrate the effectiveness of GeoReVis.
\end{abstract}

\begin{figure}[!t]
\centering\includegraphics[width=\linewidth]{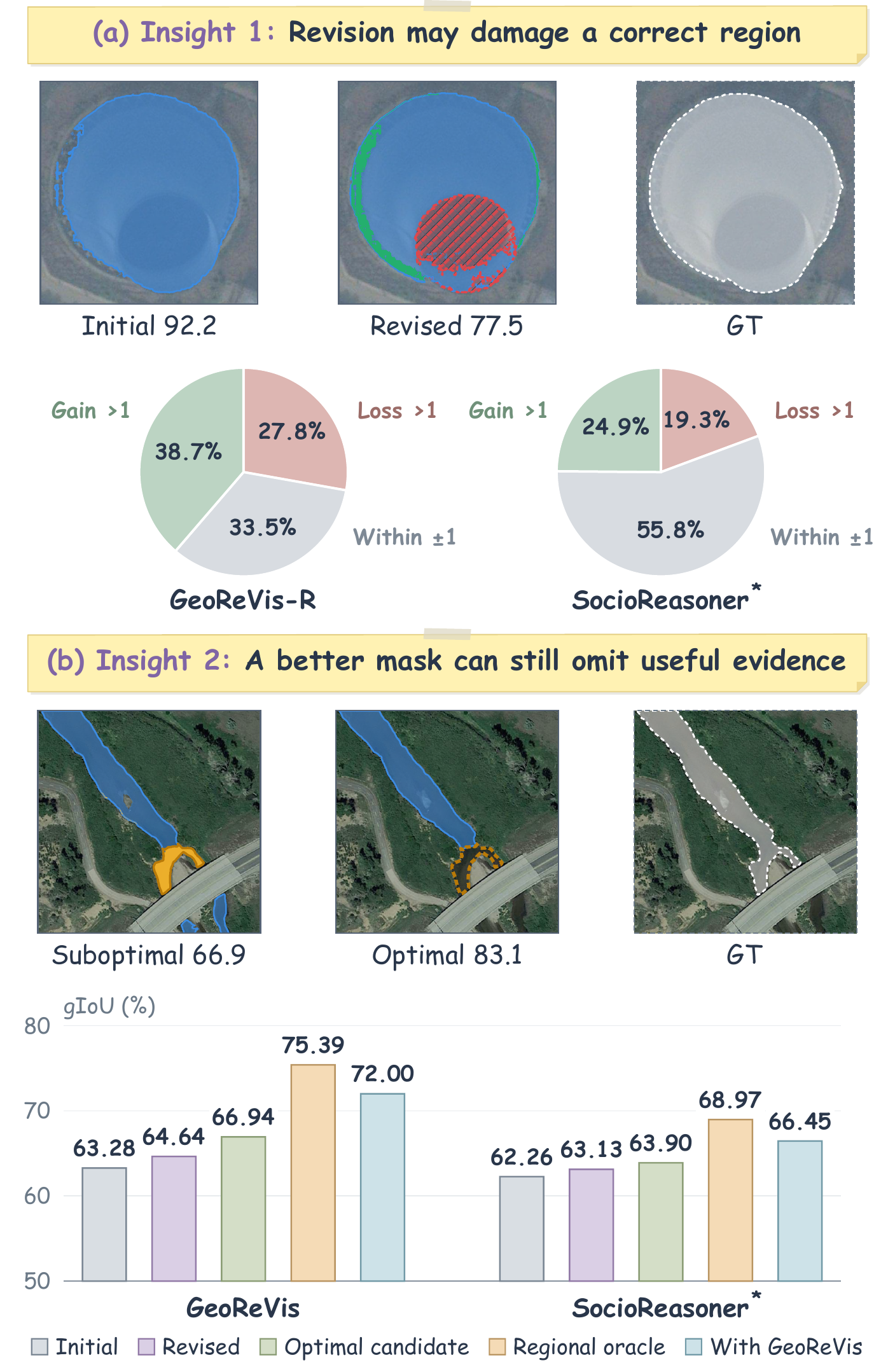}
\caption{\textbf{Analysis of Revision Degradation and Evidence Fragmentation.} \textbf{(a)} Revision repairs the tower rim (green) but damages correct support (red). Pies summarize 1,158 LaSeRS samples. \textbf{(b)} A better river mask drops a useful segment (amber). The gap between the optimal candidate and regional oracle reveals useful evidence left in suboptimal candidates, motivating Cross-Mask Evidence Consolidation. Examples use illustrative masks on real imagery.}
\label{fig:motivation}
\end{figure}

\section{Introduction}
Language-guided remote sensing segmentation identifies the pixels specified by instructions that refer to multiple targets at different granularities~\cite{segearthr2,mmr}. A prevailing paradigm uses a vision-language model (VLM) to interpret the instruction and generate segmentation prompts, which a segmentation decoder converts into masks~\cite{geopixel,segearthr2,think2segrs,rsrefseg2}. Remote sensing scenes contain densely arranged, similar-looking objects with interwoven structures and complex spatial relations~\cite{rmsin,refsegrs,geopixel}. Even an accurately located prompt may therefore fail to delimit the intended target: the predicted mask can include surrounding background or closely associated parts that fall outside the instruction's scope~\cite{rsrefseg2,think2segrs}. The existing single-pass paradigm is inherently disadvantaged in handling such errors.

Recent methods address this problem by feeding predicted masks back to the model for feedback revision~\cite{socioreasoner,iris,segagent,samveteran}, or by comparing multiple predictions to select a better candidate mask~\cite{rsagent}. These methods extend single-pass prediction through feedback and candidate selection. However, these solutions raise two subsequent questions: \textbf{Does revision necessarily improve the initial mask? Even if it does, is the best candidate mask the best segmentation the system can obtain?}

Revision can change pixels beyond its intended correction region. In Fig.~\ref{fig:motivation} (a), revision recovers the tower rim but removes a larger correct region, lowering intersection over union (IoU) from 92.2\% to 77.5\%.

We assess this \textbf{Revision Degradation} across all 1,158 LaSeRS samples using two feedback-based models: GeoReVis-R and SocioReasoner$^\ast$. GeoReVis-R is our revision-only variant of GeoReVis, while SocioReasoner$^\ast$ uses Qwen3.5-2B and frozen SAM3, trained on LaSeRS with optical imagery and no digital maps. The revision improves more than 1\% IoU on 38.7\% and 24.9\% samples. However, it decreases more than 1\% IoU on 27.8\% and 19.3\% samples at the same time. These results reveal that Revision Degradation is prevalent, even when the mean change in IoU is positive (Supplementary Sec.~A.1). Revision should not be adopted unconditionally. These observations lead to \textbf{Insight 1: Revision may damage a correct region.}

Beyond Revision Degradation, we find that the IoU of the revised mask is nearly on par with that of the initial mask on 33.51\% and 55.79\% samples. After inspection, we discover that this does not mean the revision does nothing. Gains and losses often emerge simultaneously and offset one another.

We further analyze this phenomenon in depth. For convenience, we denote the mask with the highest IoU as the optimal candidate and the other masks as suboptimal candidates. In Fig.~\ref{fig:motivation} (b), the optimal candidate correctly segments the downstream water and reaches 83.1\% IoU, yet it drops the river adjoining the bridge, 15.2\% of the target area, which a suboptimal candidate (66.9\%) had captured. At this time, selecting either whole mask leaves useful target evidence unused.

We perform examination on 1158 samples. Even assuming the optimal candidate can always be selected, the IoU remains far lower than that of the regional oracle that accommodates target evidence from all candidates (GeoReVis-R: 66.94 vs 75.39, SocioReasoner$^\ast$: 63.90 vs 68.97). We term this dispersion of correct target evidence across imperfect candidates \textbf{Evidence Fragmentation} (Supplementary Sec.~A.2). Higher whole-mask quality does not imply that all useful evidence has been retained, yielding \textbf{Insight 2: An optimal candidate mask can still omit useful evidence in suboptimal candidates.}

These insights motivate \textbf{GeoReVis}, a framework that learns \textbf{Selective Revision} and \textbf{Evidence Consolidation} for remote sensing segmentation. First, \textbf{Feedback-Guided Revision} learns fine-grained actions whose benefits have been verified through execution. Second, \textbf{Candidate Selection} compares the resulting candidate masks and retains a reference mask to reduce Revision Degradation.

Evidence Fragmentation, however, means that optimal mask selection can still discard useful target evidence. We therefore introduce \textbf{Cross-Mask Evidence Consolidation}, which retains all candidates, including rejected masks, and partitions their spatial disagreements into composable regions. Conditioned on the full instruction and image context, the model jointly determines which regions to combine, linking local decisions to the quality of the reconstructed mask. It then compares this reconstruction with the selected reference mask to determine the next round's input.

Correct regions can thus contribute independently of their source candidate, converting evidence scattered across candidates into a more correct segmentation. As shown in Fig.~\ref{fig:motivation} (b), GeoReVis can effectively mitigate Revision Degradation and Evidence Fragmentation, significantly boosting segmentation accuracy to near the regional oracle.

\begin{figure*}[t]
\centering\includegraphics[width=\textwidth]{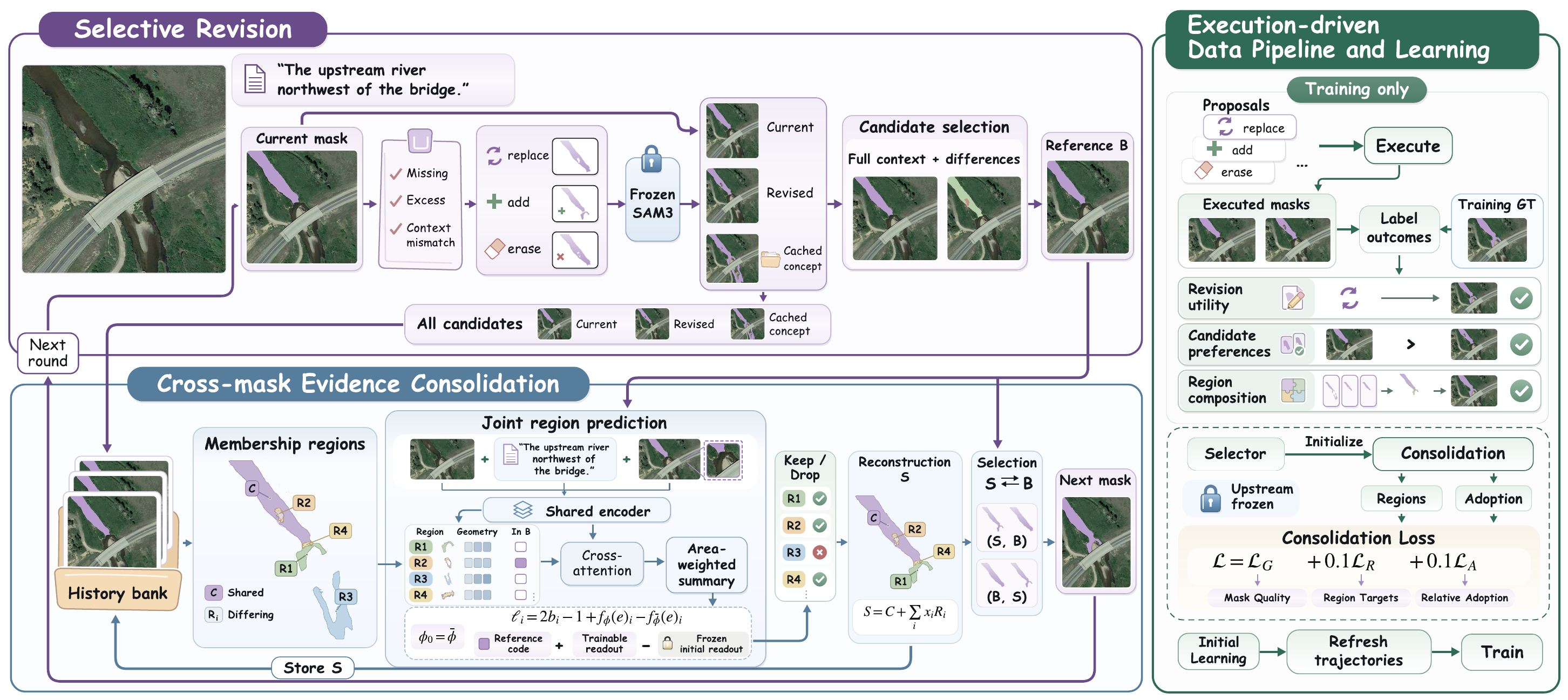}
\caption{\textbf{GeoReVis.} \textbf{Upper left:} Selective Revision uses mask feedback to generate executable corrections and compares actual candidates to select a reference $B$. \textbf{Lower left:} Cross-Mask Evidence Consolidation retains all candidates, partitions their spatial disagreements, and jointly predicts binary inclusion decisions $x_i$. The reconstruction $S$ is compared with $B$ before adoption and stored for future rounds. \textbf{Right:} the Execution-Driven Data Pipeline derives action, preference, and composition supervision from actual outputs. The example uses recorded LaSeRS candidate masks and a regional-oracle reconstruction, with four representative regions shown.}
\label{fig:method}
\vspace{-6pt}
\end{figure*}

To support GeoReVis' learning, we build an \textbf{Execution-Driven Data Pipeline} that generates fine-grained multi-round segmentation trajectories and derives supervision for revision actions, candidate selection, and region composition from the model's own outputs. We label these outputs based on ground truth and segmenter evaluation, teaching each component which decisions improve segmentation based on its actual behavior. This extends training beyond coarse whole-mask IoU supervision. We also propose \textbf{ReVis-Eval}, a benchmark protocol for measuring Revision Degradation and Evidence Fragmentation. Our contributions are:
\begin{itemize}[leftmargin=*,itemsep=2pt,topsep=3pt]
\item We reveal Revision Degradation and Evidence Fragmentation in language-guided remote sensing segmentation: revision can damage correct regions, and optimal candidates can still omit useful evidence.
\item We propose \method, which learns Selective Revision and Cross-Mask Evidence Consolidation through an Execution-Driven Data Pipeline, reducing Revision Degradation and recovering evidence from suboptimal candidates.
\item GeoReVis achieves state-of-the-art performance on five remote sensing benchmarks. We also introduce \mbox{ReVis-Eval} to assess Revision Degradation and Evidence Fragmentation, demonstrating effective feedback revision and evidence recovery.
\end{itemize}

\section{Related Work}
\paragraph{Language-guided remote sensing segmentation.}
Conventional task-specific segmenters such as RMSIN encode images and language separately~\cite{rmsin}. RSRefSeg~2 and CROSS couple contrastive VLMs with SAM to strengthen spatial grounding~\cite{rsrefseg2,cross}. Generative VLMs enable implicit reasoning~\cite{lisa}: GeoPixel and SegEarth-R2 connect language reasoning to mask decoding~\cite{geopixel,segearthr2}, while Think2Seg-RS learns spatial prompts for a frozen segmenter~\cite{think2segrs}. Beyond instruction-to-mask grounding, GeoReVis learns when to adopt corrections by supervising revision actions and candidate selection with actual segmentation outcomes.

\paragraph{Feedback revision and candidate selection in segmentation.}
Generative VLMs in SegAgent and SAM-Veteran learn mask-guided corrective prompting~\cite{segagent,samveteran}. SocioReasoner revises spatial prompts from satellite and map observations~\cite{socioreasoner}. RSAgent uses rendered mask history for tool interaction and candidate selection~\cite{rsagent}, while GuideSeg verifies proposed regions before refinement~\cite{guideseg}. FocalClick, a non-VLM interactive segmenter, preserves existing regions through local refinement and progressive merging~\cite{focalclick}. GeoReVis extends Candidate Selection to Cross-Mask Evidence Consolidation by partitioning spatial disagreements into regions, including rejected candidates' evidence. Reconstructed-mask quality supervises joint composition under the full instruction, decoupling local evidence reuse from candidate quality.

\section{GeoReVis}
\label{sec:method}
GeoReVis separates mask revision, Candidate Selection, and region composition (Fig.~\ref{fig:method}). \textbf{Selective Revision} addresses Revision Degradation, while \textbf{Cross-Mask Evidence Consolidation} recovers useful regions omitted by whole-mask selection. The \textbf{Execution-Driven Data Pipeline} supervises these decisions through their actual segmentation effects.

Given image $I$ and instruction $q$, Target Decomposition and Spatial Grounding prompt frozen SAM3~\cite{sam3} to produce initial masks $M_0$. Each target undergoes three rounds of revision, selection, and consolidation before the final masks are merged. All learned roles use Qwen3.5-2B~\cite{qwen35} with role-specific LoRA adapters~\cite{lora}. We describe one target under the complete instruction (Supplementary Secs.~B.1--B.2).

\subsection{Selective Feedback Revision Learning}
\label{sec:execution}
\paragraph{Feedback Revision targets observed segmentation errors.}
At round $t$, Feedback Assessment examines the image, instruction, current mask $M_{t-1}$, and decision history (Fig.~\ref{fig:method}, upper left). It identifies missing or excess regions and contextual mismatches. The reviser specifies an action $a$ (\emph{replace}, \emph{add}, or \emph{erase}) with boxes and signed points. SAM3 returns support $E_a$, giving
\begin{equation}
 H_a=\begin{cases}
 E_a,&\mathrm{op}(a)=\mathrm{replace},\\
 M_{t-1}\cup E_a,&\mathrm{op}(a)=\mathrm{add},\\
 M_{t-1}\setminus E_a,&\mathrm{op}(a)=\mathrm{erase}.
 \end{cases}
 \label{eq:revision}
\end{equation}
For addition and erasure, $E_a$ is clipped to the specified editing boxes. This controls the correction type and extent, enabling local repair without regenerating the mask.

\paragraph{Candidate Selection separates correction from adoption.}
Candidate Selection evaluates executed masks under the complete instruction, since plausible actions can degrade segmentation. Its candidate pool $\mathcal P_t$ includes the current mask, executed revisions, and a cached Concept Candidate generated once from a semantic prompt. This complements box-based grounding for connected structures with ambiguous spatial extent.

The selector compares full-mask and aligned difference views. Full views establish target identity and spatial relations, while differences expose added or removed regions. Pairwise comparisons with parameters $\psi$ produce the reference
\begin{equation}
 B_t=\operatorname{Select}_{\psi}(I,q,\mathcal P_t).
 \label{eq:selection}
\end{equation}
All candidates, including rejected masks, enter the History Bank $\mathcal H_t$ for later evidence reuse.

\subsection{Cross-Mask Evidence Consolidation}
\label{sec:consolidation}
\paragraph{Membership Regions make fragmented evidence composable.}
To recover a correct part from a suboptimal candidate, we partition spatial disagreements across $\mathcal H_t$. After deduplication, each pixel has a binary membership signature across the $J$ masks. Pixels with the same nonconstant signature are split into eight-connected regions $R_1,\ldots,R_K$. Their boundaries follow observed disagreements, preserving small targets and narrow structures.

Let $C_t$ denote the shared foreground and $x_i\in\{0,1\}$ indicate whether $R_i$ is retained. Using binary masks, reconstruction is
\begin{equation}
 S_t(x)=C_t+\sum_{i=1}^{K}x_iR_i.
 \label{eq:reconstruct}
\end{equation}
Every candidate, including $B_t$, is exactly representable. Additional history refines the partition without losing earlier feasible reconstructions (Supplementary Sec.~B.3). This expands whole-mask selection to complementary region composition. In Fig.~\ref{fig:method} (lower left), the partition separates bridge-adjacent target evidence from unwanted water beyond the bridge.

\paragraph{Joint Region Prediction links local choices to the whole target.}
Independent region judgments can recover target pixels while introducing enough background to reduce overall IoU. Joint Region Prediction conditions all inclusions on the complete instruction and image context. A Shared Encoder processes the image, reference overlay, and zoomed reference view. Each region pools visual features and appends geometry, relative area, and its inclusion bit $b_i$ in $B_t$.

Cross-attention~\cite{attention} connects these regional features to the shared context. An Area-Weighted Summary aggregates the resulting features $z_i$:
\begin{equation}
 \bar z=\frac{\sum_{i=1}^{K}|R_i|z_i}{\sum_{i=1}^{K}|R_i|}.
 \label{eq:summary}
\end{equation}
Concatenating $\bar z$ with each $z_i$ lets the readout evaluate a region in relation to the complete composition. Area weighting accounts for the different spatial impacts of small target parts and large background regions.

The readout learns changes around the selected reference:
\begin{equation}
 \ell_i=2b_i-1+f_\phi(e)_i-f_{\bar\phi}(e)_i,
 \qquad \phi_0=\bar\phi,
 \label{eq:identity}
\end{equation}
where $e$ collects the shared context and regional features. The trainable readout $f_\phi$ and its frozen initial copy $f_{\bar\phi}$ receive the same features. Their outputs initially cancel, reproducing $B_t$ so the model learns regional changes from the selected mask. Thresholding the logits yields $x$, and Eq.~\ref{eq:reconstruct} forms $S_t$ without another segmenter call (Supplementary Sec.~B.4).

\paragraph{Relative Adoption selects the next mask and preserves alternatives.}
The consolidation adapter compares $S_t$ with $B_t$ in both input orders. Each margin is the reconstruction preference logit minus the reference preference logit. Averaging the two margins gives $m_t$ and reduces presentation-order bias~\cite{positionbias}. The next round receives
\begin{equation}
 M_t=\begin{cases}S_t,&m_t>0,\\B_t,&m_t\leq0.\end{cases}
 \label{eq:adoption}
\end{equation}
This applies the principle of Selective Revision to consolidation itself. Regardless of adoption, $S_t$ joins the History Bank for later composition. Supplementary Sec.~B.6 relates adoption errors to segmentation quality.

\subsection{Execution-Driven Data Pipeline}
\label{sec:joint}
\paragraph{Actual executions supervise each decision.}
We first train Target Decomposition and Spatial Grounding on benchmark annotations to establish Initial Segmentation. The pipeline in Fig.~\ref{fig:method} (right) generates training states through the same interfaces as inference. Teacher proposals, proposal-seeded search, and rules supply candidate actions. With training mask $G$, their utility is
\begin{equation}
\Delta_a=\IoU(H_a,G)-\IoU(M_{t-1},G).
\label{eq:action_utility}
\end{equation}
Actions with positive utility supervise Feedback Revision. Actual mask pairs receive IoU-based preference labels for Candidate Selection. These labels distinguish a convincing correction instruction from one that improves the executed segmentation (Supplementary Sec.~C.1).

Multi-round trajectories supply candidate banks for consolidation. The bank determines the regions before GT assigns composition and adoption targets. Feedback-conditioned revision and multi-round selection training additionally expose the model to its own observed errors and earlier decisions, addressing the training--inference state mismatch in sequential prediction~\cite{dagger,gkd}.

\paragraph{Joint learning optimizes complete-mask improvement.}
The consolidation adapter inherits the multi-round selector's parameters, transferring its learned comparative representation. Upstream roles and SAM3 remain frozen. For $N$ target states, let $I_n$ and $U_n$ denote the intersection and union of reconstruction $n$ with GT. Following the principle of optimizing the segmentation metric directly~\cite{lovasz}, the Mask Quality loss combines target-averaged and pooled IoU:
\begin{equation}
 \mathcal L_G=\tfrac12\left(1-\tfrac1N\sum_n\tfrac{I_n}{U_n}\right)
 +\tfrac12\left(1-\tfrac{\sum_n I_n}{\sum_n U_n}\right).
 \label{eq:global}
\end{equation}
The first term preserves small targets' influence, while the second accounts for their pixel contribution. Hard reconstructions are scored in the forward pass, with straight-through gradients~\cite{ste} through relaxed inclusions. Thus, region choices are optimized for their joint segmentation effect.

A weighted region loss $\mathcal L_R$ supervises the exact IoU-optimal inclusion code within each bank. A Relative Adoption loss $\mathcal L_A$ learns reconstruction-versus-reference preferences, weighted by their realized IoU differences. It also distills the original selector's judgments when they agree with GT ordering. The shared consolidation adapter and regional readout are trained with the Consolidation Loss
\begin{equation}
 \mathcal L=\mathcal L_G+0.1\mathcal L_R+0.1\mathcal L_A.
 \label{eq:joint}
\end{equation}
The auxiliary terms teach regional evidence use and comparative judgment. Initial learning uses the collected trajectories. We then refresh trajectories through re-execution and continue training on the augmented states, adapting consolidation to its own decisions. Label derivations, exact losses, and training schedules appear in Supplementary Secs.~B.5 and C.2--C.3.

\section{ReVis-Eval}
\label{sec:revis}
ReVis-Eval is a benchmark protocol built on LaSeRS~\cite{segearthr2} with two evaluations. \emph{Revision Degradation Evaluation} measures feedback-driven improvement and preservation of correct regions. \emph{Evidence Fragmentation Evaluation} measures whether consolidating evidence across candidates improves on the optimal candidate.

\paragraph{Revision Degradation Evaluation.}
We use the segmentation pipelines of GeoReVis, VisionReasoner$^\ddagger$, and SocioReasoner$^\ddagger$ without revision to generate one initial mask per model for each of the 1,158 LaSeRS samples. Filtering out masks with IoU above 0.95, which have limited room for improvement, leaves 3,201 initial masks.

During evaluation, each method receives the same image, instruction, and initial mask for each of the 3,201 items and performs three revision rounds. After that, we report four metrics:
\begin{enumerate}[label=\textbf{(\arabic*)},leftmargin=*,itemsep=2pt,topsep=3pt,parsep=0pt]
\item \textbf{IoU Gain ($\Delta$IoU)}: Mean IoU difference between final and initial masks. Higher $\Delta$IoU indicates greater improvement through revision.
\item \textbf{Feedback Gain (FG)}: Mean final-IoU difference between paired runs with and without mask overlays at the same budget. Higher FG indicates more effective use of segmentation feedback.
\item \textbf{Degradation Rate (DR)}: Proportion of initial masks whose IoU drops by more than 0.01 after revision. Lower DR indicates less frequent Revision Degradation.
\item \textbf{Correct-Pixel Preservation (CP)}: Mean fraction of initially correct target pixels retained in the final mask. Higher CP indicates better preservation of correct regions.
\end{enumerate}

\paragraph{Evidence Fragmentation Evaluation.}
We run GeoReVis without evidence consolidation, SAM-Veteran$^\ast$, and SocioReasoner$^\ast$ for three revision rounds on 1,158 LaSeRS samples and collect their trajectories. We retain all candidate masks from each trajectory and take its final output as the reference mask $B$, yielding 3,474 banks with \mbox{4-10} masks each.

During evaluation, tested methods receive the same image, instruction, and candidate bank, and produce a final mask without additional segmenter calls. Alongside final IoU, we report four metrics:
\begin{enumerate}[label=\textbf{(\arabic*)},leftmargin=*,itemsep=2pt,topsep=3pt,parsep=0pt]
\item \textbf{Best-Whole Gap (BWG)}: Mean IoU difference between the final mask and the candidate with the highest GT IoU. Positive BWG indicates improvement beyond optimal whole-mask selection.
\item \textbf{BWG on High-Potential Banks (BWG$_P$)}: BWG averaged over 812 banks whose regional oracle exceeds the optimal candidate by more than 0.05 IoU. Higher BWG$_P$ indicates better use of available consolidation potential.
\item \textbf{Historical Recall (HR)}: Mean fraction of target pixels present in the bank but absent from $B$ that are recovered in the final mask. Higher HR indicates more complete recovery of fragmented evidence.
\item \textbf{Added Precision (AP)}: Mean fraction of pixels added to $B$ that belong to the target. Higher AP indicates more precise evidence recovery.
\end{enumerate}
Construction details and metric formulas appear in Supplementary Secs.~D.2--D.3.

\section{Experiments}
\label{sec:experiments}
\begin{table*}[t]
\centering\caption{\textbf{Comparison on LaSeRS.} We report gIoU for each of the nine dimensions. The last two columns report the official macro gIoU and cIoU (\%) across these dimensions. $^\ddagger$: retrained from released code on the same training split. $^\ast$: trained with Qwen3.5-2B and frozen SAM3 on LaSeRS. Best and second-best results are in bold and underlined.}\label{tab:lasers}
\setlength{\tabcolsep}{3.6pt}\small
\resizebox{\linewidth}{!}{\begin{tabular}{lll|ccccccccc|cc}\toprule
\multicolumn{1}{l}{\multirow{2}{*}[-0.4ex]{\tablehead{Method}}} & \multicolumn{1}{l}{\multirow{2}{*}[-0.4ex]{\tablehead{Backbone}}} & \multicolumn{1}{l|}{\multirow{2}{*}[-0.4ex]{\tablehead{Segmenter}}} & \multicolumn{3}{c}{\tablehead{Granularity}} & \multicolumn{2}{c}{\tablehead{Multiplicity}} & \multicolumn{2}{c}{\tablehead{Reasoning}} & \multicolumn{2}{c|}{\tablehead{Length}} & \multicolumn{2}{c}{\tablehead{Overall}}\\
\cmidrule(lr){4-6}\cmidrule(lr){7-8}\cmidrule(lr){9-10}\cmidrule(lr){11-12}\cmidrule(lr){13-14}
& & & \tablehead{Sem.} & \tablehead{Inst.} & \tablehead{Part} & \tablehead{Single} & \tablehead{Multi} & \tablehead{Expl.} & \tablehead{Impl.} & \tablehead{Short} & \tablehead{Long} & \tablehead{gIoU} & \tablehead{cIoU}\\\midrule
\multicolumn{14}{c}{\textbf{\textit{Single-pass Segmentation Models}}}\\\midrule
LISA~\cite{lisa} & LLaVA-7B & SAM & 26.40 & 20.50 & 16.10 & 37.30 & 18.20 & 27.10 & 21.50 & 34.10 & 38.40 & 26.60 & 25.10\\
M$^2$A~\cite{mmr} & LLaVA-7B & SAM & 30.10 & 23.00 & 18.60 & 45.40 & 20.90 & 35.80 & 23.30 & 35.80 & 41.50 & 30.50 & 29.30\\
GLaMM-ft~\cite{glamm} & Vicuna-7B & SAM & 44.80 & 41.20 & 32.60 & 47.30 & 32.20 & 59.10 & 42.60 & 50.40 & 42.60 & 43.60 & 48.30\\
GeoPixel~\cite{geopixel} & IXC-2.5-7B & SAM2 & 51.40 & 44.10 & 43.90 & 55.00 & \underline{49.20} & 66.50 & 41.10 & 51.10 & 51.40 & 50.40 & 55.20\\
Seg-Zero$^\ddagger$~\cite{segzero} & Qwen2.5-VL-7B & SAM2 & 47.37 & 60.60 & 57.84 & 48.64 & 31.16 & 66.76 & 43.16 & 60.23 & 59.18 & 52.77 & 58.50\\
VisionReasoner$^\ddagger$~\cite{visionreasoner} & Qwen2.5-VL-7B & SAM2 & 48.98 & 62.05 & 59.52 & 49.15 & 33.31 & 67.22 & 44.31 & 60.62 & 61.12 & 54.03 & 59.80\\
Think2Seg-RS$^\ddagger$~\cite{think2segrs} & Qwen2.5-VL-7B & SAM2 & 52.08 & 65.37 & 60.85 & 52.11 & 37.72 & 69.78 & 46.70 & 64.05 & 63.88 & 56.95 & 65.52\\
SegEarth-R2~\cite{segearthr2} & Mipha-3B & Mask2Former & 60.20 & 65.40 & 64.80 & 55.10 & 38.30 & \underline{78.40} & 42.80 & 60.20 & 50.10 & 57.20 & 67.90\\
VisionReasoner$^\ast$ & Qwen3.5-2B & SAM3 & 60.71 & 70.62 & 67.42 & 59.85 & 45.48 & 75.78 & 55.48 & 69.58 & 70.55 & 63.94 & 64.38\\
GeoReason$^\ddagger$~\cite{georeason} & Qwen3-VL-4B & SAM2 & \underline{62.18} & \underline{70.96} & \underline{69.14} & \underline{60.62} & 46.86 & 75.48 & \underline{57.36} & 69.24 & \underline{72.81} & \underline{64.96} & 68.72\\
\midrule
\multicolumn{14}{c}{\textbf{\textit{Feedback-revision Segmentation Models}}}\\\midrule
SAM~3 Agent~\cite{sam3} & Qwen2.5-VL-72B & SAM3 & 37.50 & 52.49 & 48.33 & 38.34 & 20.08 & 58.44 & 32.32 & 51.33 & 50.51 & 43.26 & 46.64\\
SegAgent$^\ddagger$~\cite{segagent} & Qwen-VL-7B & SAM & 50.37 & 63.32 & 60.64 & 51.93 & 33.64 & 68.18 & 45.88 & 61.81 & 62.12 & 55.32 & 62.19\\
SAM-Veteran$^\ddagger$~\cite{samveteran} & Qwen2.5-VL-7B & SAM2 & 53.34 & 66.12 & 61.62 & 53.32 & 38.19 & 70.93 & 48.15 & 63.77 & 63.23 & 57.63 & 63.93\\
SocioReasoner$^\ddagger$~\cite{socioreasoner} & Qwen2.5-VL-3B & SAM2 & 55.73 & 67.04 & 63.99 & 55.13 & 39.85 & 70.79 & 49.77 & 65.69 & 65.99 & 59.33 & 67.34\\
SAM-Veteran$^\ast$ & Qwen3.5-2B & SAM3 & 59.39 & 69.20 & 66.78 & 58.48 & 44.26 & 74.52 & 54.98 & 67.75 & 69.03 & 62.71 & 68.42\\
SocioReasoner$^\ast$ & Qwen3.5-2B & SAM3 & 58.86 & 69.78 & 67.70 & 59.86 & 45.71 & 74.29 & 54.16 & \underline{69.61} & 68.20 & 63.13 & \underline{68.97}\\
\rowcolor{paperteal!9}\textbf{GeoReVis (ours)} & Qwen3.5-2B & SAM3 & \textbf{68.11} & \textbf{78.56} & \textbf{75.95} & \textbf{68.60} & \textbf{54.42} & \textbf{83.29} & \textbf{63.73} & \textbf{77.63} & \textbf{77.73} & \textbf{72.00} & \textbf{77.01}\\
\bottomrule
\end{tabular}}\end{table*}

\subsection{Experimental Setup}
\paragraph{Benchmarks and Datasets.}
We evaluate on five remote sensing segmentation benchmarks. LaSeRS~\cite{segearthr2} is the main benchmark, with 1,158 test queries covering nine dimensions across target granularity, multiplicity, reasoning, and instruction length. EarthReason~\cite{segearthr1} evaluates reasoning segmentation, while RRSIS-D~\cite{rmsin}, RefSegRS~\cite{refsegrs}, and RISBench~\cite{crobim} evaluate referring segmentation. GeoReVis is trained separately on each benchmark's corresponding training split. ReVis-Eval uses held-out LaSeRS scenes to assess feedback revision and Cross-Mask Evidence Consolidation under the controlled inputs specified in Sec.~\ref{sec:revis}.

\paragraph{Evaluation Metrics.}
Following the benchmark protocols and prior studies~\cite{segearthr2,unigeoseg}, we use gIoU and cIoU as evaluation metrics in all experiments. We report performance across nine key dimensions on LaSeRS~\cite{segearthr2}. ReVis-Eval additionally includes Revision Degradation Evaluation and Evidence Fragmentation Evaluation. Its metric definitions are given in Sec.~\ref{sec:revis}.

\paragraph{Baselines.}
As shown in Tables~\ref{tab:lasers}, \ref{tab:reasoning}, and~\ref{tab:referring}, we conduct a comprehensive comparison with existing SOTA single-pass and feedback-revision segmentation models. All feedback-revision models are evaluated for three revision rounds. EarthReason comparisons focus on generative VLM-based reasoning segmenters. Referring benchmarks additionally include task-specific models designed for explicit expressions. For fairness, we report published results on each benchmark when available. Otherwise, we retrain released implementations on the same settings and mark these reproductions with $^\ddagger$.

To control for backbone and segmenter differences, we additionally train VisionReasoner, SAM-Veteran, and SocioReasoner with Qwen3.5-2B and frozen SAM3, denoted by $^\ast$. For SocioReasoner, we provide only the satellite image. Retraining settings and model-specific adaptations appear in Supplementary Sec.~D.1.

\paragraph{Implementation Details.}
We use Qwen3.5-2B and frozen SAM3 as a backbone for GeoReVis. We train rank-16 LoRA adapters for target decomposition, grounding, feedback assessment, revision, candidate selection, and evidence consolidation. We use AdamW~\cite{adamw} with effective batch sizes of 16. Consolidation training uses four A6000 GPUs, with first-pass learning rates of $5\times10^{-6}$ and $3\times10^{-4}$ for the adapter and regional readout, respectively.

At inference, GeoReVis performs three revision rounds, each with Candidate Selection and Evidence Consolidation. The readout architecture and training schedules appear in Supplementary Secs.~B.4 and C.3.

\subsection{Main Results}
\begin{table}[t]\centering
\caption{\textbf{Reasoning segmentation on EarthReason} (gIoU/cIoU, \%). Avg. is the arithmetic mean of the four metrics.}\label{tab:reasoning}
\footnotesize\setlength{\tabcolsep}{2pt}
\begin{tabularx}{\linewidth}{L{2.3}*{5}{C{0.74}}}\toprule
\multirow{2}{*}[-0.4ex]{\tablehead{Method}} & \multicolumn{2}{c}{\tablehead{Val}} & \multicolumn{2}{c}{\tablehead{Test}} & \multirow{2}{*}[-0.4ex]{\tablehead{Avg.}}\\
\cmidrule(lr){2-3}\cmidrule(lr){4-5}
& \tablehead{gIoU} & \tablehead{cIoU} & \tablehead{gIoU} & \tablehead{cIoU} & \\\midrule
LISA~\cite{lisa} & 61.04 & 57.39 & 60.88 & 59.10 & 59.60\\
RemoteReasoner~\cite{remotereasoner} & 69.02 & 67.80 & 70.96 & 69.13 & 69.23\\
SegEarth-R2~\cite{segearthr2} & 72.30 & 68.10 & 73.50 & 69.50 & 70.85\\
UniGeoSeg~\cite{unigeoseg} & 72.54 & 73.30 & 73.08 & 74.61 & 73.38\\
Think2Seg-RS~\cite{think2segrs} & 72.16 & 72.87 & 73.36 & 75.60 & 73.50\\
VisionReasoner$^\ast$~\cite{visionreasoner} & 74.36 & 75.71 & 74.57 & 75.29 & 74.98\\
SocioReasoner$^\ast$~\cite{socioreasoner} & 74.07 & 74.91 & 75.32 & 76.18 & 75.12\\
GeoReason~\cite{georeason} & \underline{77.61} & \underline{77.63} & \underline{78.61} & \underline{80.60} & \underline{78.61}\\
\rowcolor{paperteal!9}\textbf{GeoReVis (ours)} & \textbf{77.63} & \textbf{79.35} & \textbf{79.28} & \textbf{81.17} & \textbf{79.36}\\
\bottomrule\end{tabularx}\end{table}

\begin{table}[t]\centering
\caption{\textbf{Referring segmentation} (test gIoU, \%). Each model uses the corresponding benchmark's training split. Additional results and cIoU metrics are in the supplement.}\label{tab:referring}
\footnotesize\setlength{\tabcolsep}{2.5pt}
\begin{tabularx}{\linewidth}{L{1.6}*{3}{C{0.8}}}\toprule
\tablehead{Method} & \tablehead{RRSIS-D} & \tablehead{RefSegRS} & \tablehead{RISBench}\\\midrule
\multicolumn{4}{c}{\textbf{\textit{Referring Segmentation Specialists}}}\\
\midrule
RS2-SAM 2$^\ddagger$~\cite{rs2sam2} & 66.72 & 73.90 & 68.55\\
RSRefSeg 2~\cite{rsrefseg2} & 69.17 & 77.39 & \underline{72.57}\\
CROSS$^\ddagger$~\cite{cross} & 68.92 & \underline{79.51} & 70.86\\
\midrule
\multicolumn{4}{c}{\textbf{\textit{Generative VLM-Based Models}}}\\
\midrule
SocioReasoner$^\ast$~\cite{socioreasoner} & 64.61 & 71.94 & 67.62\\
SegEarth-R2~\cite{segearthr2} & 67.90 & 74.80 & 70.50\\
GeoReason$^\ddagger$~\cite{georeason} & 67.32 & 76.38 & 71.36\\
UniGeoSeg$^\ddagger$~\cite{unigeoseg} & \underline{69.25} & 76.06 & 70.94\\
\rowcolor{paperteal!9}\textbf{GeoReVis (ours)} & \textbf{70.35} & \textbf{80.72} & \textbf{74.27}\\
\bottomrule\end{tabularx}\end{table}

\paragraph{Results on LaSeRS.}
LaSeRS is the most challenging RS segmentation benchmark~\cite{segearthr2}. GeoReVis achieves 72.00\% gIoU and 77.01\% cIoU, leading all nine dimensions in Table~\ref{tab:lasers}. It improves gIoU over GeoReason, the SOTA single-pass segmentation model, by 7.04\% and over SegEarth-R2, designed for multi-target and multi-granularity segmentation, by 14.80\%. GeoReVis surpasses the backbone-matched VisionReasoner$^\ast$ and feedback-revision model SocioReasoner$^\ast$ by 8.06\% and 8.87\% gIoU, respectively. These results establish GeoReVis's broad advantage across hierarchical granularity, target multiplicity, reasoning requirements, and linguistic variability in complex remote sensing scenes.

\paragraph{Results on EarthReason.}
GeoReason emphasizes instruction understanding and spatial grounding~\cite{georeason}. GeoReVis leads all four EarthReason metrics, averaging 79.36\% (Table~\ref{tab:reasoning}). These results demonstrate more accurate interpretation of diverse implicit instructions and segmentation of the intended targets. Despite annotation noise in execution-driven supervision, some predictions delineate targets even more accurately than GT masks (Supplementary Sec.~E.7).

\paragraph{Results on Referring Segmentation Benchmarks.}
GeoReVis also achieves the highest gIoU on all three referring benchmarks (Table~\ref{tab:referring}). It surpasses UniGeoSeg on RRSIS-D by 1.10\%, CROSS on RefSegRS by 1.21\%, and RSRefSeg~2 on RISBench by 1.70\%. Thus, GeoReVis effectively delineates explicitly specified targets as well as those described through complex implicit instructions.

\paragraph{Overall.}
With only 2B parameters, GeoReVis leads all five benchmarks. Its advantage is largest on LaSeRS, which additionally stresses part-level delineation and multi-target composition~\cite{segearthr2}. These tasks expose the two problems identified in our insights: revision may damage a correct region, and an optimal candidate mask can still omit useful evidence in suboptimal candidates. GeoReVis effectively mitigates Revision Degradation and Evidence Fragmentation (Supplementary Sec.~E.1). Backbone choice has little effect on performance (Supplementary Sec.~E.8).

\subsection{Results On ReVis-Eval}
\begin{table}[t]\centering
\caption{\textbf{Revision Degradation Evaluation on ReVis-Eval.} R: revision only. RS: revision and Candidate Selection, including the concept candidate. Metrics are defined in Sec.~\ref{sec:revis}.}\label{tab:revis_a}
\small\setlength{\tabcolsep}{2pt}\renewcommand{\arraystretch}{1.08}
\begin{tabularx}{\linewidth}{L{1.8}*{4}{C{0.8}}}\toprule
\tablehead{Method} & \tablehead{$\Delta$IoU$\uparrow$} & \tablehead{FG$\uparrow$} & \tablehead{DR$\downarrow$} & \tablehead{CP$\uparrow$}\\\midrule
SegAgent$^\ast$ & $-$0.36 & +0.43 & 30.68 & 87.36\\
SAM-Veteran$^\ast$ & +0.86 & +0.71 & 19.43 & 92.18\\
SocioReasoner$^\ast$ & +1.24 & +0.86 & 17.90 & 93.04\\
GeoReVis-R & +0.39 & +1.12 & 31.15 & 88.27\\
GeoReVis-RS & +4.94 & +1.95 & 7.72 & 95.37\\
\rowcolor{paperteal!9}\textbf{GeoReVis} & \textbf{+5.28} & \textbf{+2.08} & \textbf{6.59} & \textbf{96.12}\\
\bottomrule\end{tabularx}
\end{table}

\begin{table}[t]\centering
\caption{\textbf{Evidence Fragmentation Evaluation on ReVis-Eval.} Metrics are defined in Sec.~\ref{sec:revis}.}\label{tab:revis_b}
\small\setlength{\tabcolsep}{1.7pt}\renewcommand{\arraystretch}{1.08}
\begin{tabularx}{\linewidth}{>{\raggedright\arraybackslash}Xccccc}\toprule
\tablehead{Method} & \tablehead{IoU$\uparrow$} & \tablehead{BWG$\uparrow$} & \tablehead{BWG$_P\uparrow$} & \tablehead{HR$\uparrow$} & \tablehead{AP$\uparrow$}\\\midrule
Union & 60.26 & $-$9.11 & $-$3.37 & 100.00 & 35.61\\
Majority Vote & 67.65 & $-$1.72 & $-$0.93 & 21.94 & 71.36\\
Whole-Mask Selection & 68.74 & $-$0.63 & $-$1.46 & 19.47 & 68.15\\
Independent Regions & 68.66 & $-$0.71 & +1.11 & 38.63 & 66.29\\
Current Round & 69.60 & +0.23 & +1.72 & 41.35 & 76.82\\
\rowcolor{paperteal!9}\textbf{GeoReVis} & \textbf{70.41} & \textbf{+1.04} & \textbf{+2.61} & \textbf{48.17} & \textbf{79.54}\\
\midrule
Optimal Candidate & 69.37 & 0.00 & 0.00 & -- & --\\
Regional Oracle & 75.76 & +6.39 & +15.12 & -- & --\\
\bottomrule\end{tabularx}
\end{table}

\paragraph{Revision Degradation Evaluation.}
GeoReVis leads all four metrics in Table~\ref{tab:revis_a}. Compared with SocioReasoner$^\ast$, IoU Gain increases from 1.24\% to 5.28\%, while Degradation Rate decreases from 17.90\% to 6.59\%. GeoReVis-R obtains positive Feedback Gain (+1.12\%) but still degrades 31.15\% of initial masks. Adding Candidate Selection and the Concept Candidate reduces this rate to 7.72\%. Full GeoReVis preserves 96.12\% of initially correct pixels. These results demonstrate effective feedback revision with reduced Revision Degradation and better preservation of correct regions.

\paragraph{Evidence Fragmentation Evaluation.}
GeoReVis reaches 70.41\% IoU, surpassing the optimal candidate at 69.37\% (Table~\ref{tab:revis_b}). This gain rises to 2.61\% on high-potential banks, using the same candidate masks without additional segmenter calls. Historical Recall reaches 48.17\%, with 79.54\% Added Precision compared with 35.61\% for Union. These results show that Cross-Mask Evidence Consolidation recovers useful regions from suboptimal candidates and converts fragmented evidence into improved segmentation. Supplementary Secs.~E.5--E.6 provide detailed process results and visual evidence.

\subsection{Ablation Studies}
\begin{table}[t]\centering
\caption{\textbf{Component Ablation on LaSeRS.} U.R.: Unverified Revision. V.R.: Verified Revision. C.S.: Candidate Selection. C.C.: Concept Candidate. M.F.: Multi-Round Feedback. E.C.: Evidence Consolidation. U.R. and V.R. are alternative revisions trained on unverified and execution-verified actions. The first row is Initial Segmentation, and the last is full GeoReVis.}\label{tab:ablation}
\footnotesize\setlength{\tabcolsep}{1.7pt}\renewcommand{\arraystretch}{1.10}
\begin{tabularx}{\linewidth}{*{6}{C{0.78}}*{4}{C{1.33}}}\toprule
\multicolumn{6}{c}{\tablehead{Components}} & \multicolumn{4}{c}{\tablehead{LaSeRS}}\\
\cmidrule(lr){1-6}\cmidrule(lr){7-10}
\tablehead{U.R.} & \tablehead{V.R.} & \tablehead{C.S.} & \tablehead{C.C.} & \tablehead{M.F.} & \tablehead{E.C.} & \tablehead{gIoU} & \tablehead{cIoU} & \tablehead{DR$\downarrow$} & \tablehead{OR$\downarrow$}\\\midrule
$\times$ & $\times$ & $\times$ & $\times$ & $\times$ & $\times$ & 63.28 & 62.85 & -- & --\\
$\checkmark$ & $\times$ & $\times$ & $\times$ & $\times$ & $\times$ & 62.71 & 62.93 & 32.47 & 37.05\\
$\times$ & $\checkmark$ & $\times$ & $\times$ & $\times$ & $\times$ & 64.64 & 66.65 & 27.81 & 36.44\\
$\times$ & $\checkmark$ & $\checkmark$ & $\times$ & $\times$ & $\times$ & 66.58 & 70.62 & 10.36 & 33.16\\
$\times$ & $\checkmark$ & $\checkmark$ & $\checkmark$ & $\times$ & $\times$ & 67.90 & 72.81 & 8.72 & 35.92\\
$\times$ & $\checkmark$ & $\checkmark$ & $\checkmark$ & $\checkmark$ & $\times$ & 69.19 & 73.43 & 7.34 & 38.69\\
\rowcolor{paperteal!9}$\times$ & $\checkmark$ & $\checkmark$ & $\checkmark$ & $\checkmark$ & $\checkmark$ & \textbf{72.00} & \textbf{77.01} & \textbf{6.13} & \textbf{27.46}\\
\bottomrule\end{tabularx}
\end{table}

\paragraph{Component Ablation.}
Table~\ref{tab:ablation} shows complementary component gains. Verified Revision improves Initial Segmentation from 63.28\% to 64.64\% gIoU, whereas Unverified Revision lowers it to 62.71\%. Candidate Selection adds 1.94\% gIoU and reduces Degradation Rate from 27.81\% to 10.36\%. The concept candidate and multi-round feedback add 1.32\% and 1.29\% gIoU, respectively. These gains demonstrate the benefits of execution-verified supervision and selective adoption.

Despite higher gIoU, multi-round feedback increases the Omitted-Evidence Rate from 35.92\% to 38.69\%. Cross-Mask Evidence Consolidation lowers it to 27.46\% and adds 2.81\% gIoU and 3.58\% cIoU. The gIoU gain reaches 4.46\% on fragmentation-heavy queries (Supplementary Table~7), demonstrating effective evidence recovery. Our method also surpasses Independent Regions, which removes only cross-region aggregation, and general aggregation strategies (Supplementary Sec.~E.3).

\paragraph{Execution-Driven Data Pipeline.}
Our pipeline supervises revision actions, candidate selection, and region composition from actual segmentation outcomes. Execution-verified labels improve single-round gIoU by 1.20\% over unverified teacher supervision and 0.86\% over synthetic pairs. Joint quality, region, and adoption supervision also benefits consolidation (Supplementary Sec.~E.2). The pipeline thus links each learned decision to its segmentation outcome.

\paragraph{Revision Rounds and Computational Cost.}
GeoReVis reaches 68.13\%, 70.26\%, 72.00\%, and 72.06\% gIoU with one to four rounds, respectively. Two rounds outperform all Table~\ref{tab:lasers} baselines at 3.98 seconds per query on four GPUs, versus 5.41 for SAM-Veteran and 6.68 for SegAgent. We use three rounds at 5.99 seconds per query, as a fourth adds little accuracy. Lightweight role adapters retain a 2B-scale VLM per call, while shared visual features reduce multi-stage overhead (Supplementary Sec.~E.4).

\section{Conclusion}
GeoReVis addresses Revision Degradation and Evidence Fragmentation through Selective Revision and Cross-Mask Evidence Consolidation. Results on five benchmarks and ReVis-Eval demonstrate superior segmentation, better preservation of correct regions, and recovery of evidence from suboptimal candidates.

\section*{LLM Usage Statement}
We used OpenAI's ChatGPT and Anthropic's Claude as assistive tools for manuscript editing, language polishing, grammar checking, code refinement and checking, figure and table refinement, and data organization. No LLMs or AI tools were used to formulate research ideas, hypotheses, claims, or conclusions, or to design experiments or analyze results. We additionally used MiniMax-M3.1 within the Execution-Driven Data Pipeline to improve training data quality. All related content was reviewed and approved by the authors. All research decisions remained with the authors, who verified the submitted content and take full responsibility for the entire manuscript, including all AI-assisted content.
{\small\bibliographystyle{ieeenat_fullname}\bibliography{references}}

@inproceedings{geopixel,
  author = {Shabbir, Akashah and Zumri, Mohammed and Bennamoun, Mohammed and Khan, Fahad Shahbaz and Khan, Salman},
  booktitle = {Proceedings of the 42nd International Conference on Machine Learning},
  pages = {54095--54111},
  publisher = {PMLR},
  title = {{{G}eo{P}ixel: Pixel Grounding Large Multimodal Model in Remote Sensing}},
  url = {https://proceedings.mlr.press/v267/shabbir25a.html},
  volume = {267},
  year = {2025}
}

@inproceedings{segearthr2,
  author = {Xin, Zepeng and Li, Kaiyu and Chen, Luodi and Li, Wanchen and Xiao, Yuchen and Qiao, Hui and Zhang, Weizhan and Meng, Deyu and Cao, Xiangyong},
  booktitle = {Proceedings of the IEEE/CVF Conference on Computer Vision and Pattern Recognition (CVPR)},
  pages = {13199--13210},
  title = {{SegEarth-R2: Towards Comprehensive Language-guided Segmentation for Remote Sensing Images}},
  url = {https://openaccess.thecvf.com/content/CVPR2026/html/Xin_SegEarth-R2_Towards_Comprehensive_Language-guided_Segmentation_for_Remote_Sensing_Images_CVPR_2026_paper.html},
  year = {2026}
}

@inproceedings{unigeoseg,
  author = {Ni, Shuo and Wang, Di and Chen, He and Guo, Haonan and Zhang, Ning and Zhang, Jing},
  booktitle = {Proceedings of the IEEE/CVF Conference on Computer Vision and Pattern Recognition (CVPR)},
  pages = {34865--34876},
  title = {{UniGeoSeg: Towards Unified Open-World Segmentation for Geospatial Scenes}},
  url = {https://openaccess.thecvf.com/content/CVPR2026/html/Ni_UniGeoSeg_Towards_Unified_Open-World_Segmentation_for_Geospatial_Scenes_CVPR_2026_paper.html},
  year = {2026}
}

@inproceedings{rmsin,
  author = {Liu, Sihan and Ma, Yiwei and Zhang, Xiaoqing and Wang, Haowei and Ji, Jiayi and Sun, Xiaoshuai and Ji, Rongrong},
  booktitle = {Proceedings of the IEEE/CVF Conference on Computer Vision and Pattern Recognition (CVPR)},
  doi = {10.1109/CVPR52733.2024.02517},
  pages = {26648--26658},
  title = {{Rotated Multi-Scale Interaction Network for Referring Remote Sensing Image Segmentation}},
  url = {https://doi.org/10.1109/CVPR52733.2024.02517},
  year = {2024}
}

@inproceedings{sam3,
  author = {Carion, Nicolas and Gustafson, Laura and Hu, Yuan-Ting and Debnath, Shoubhik and Hu, Ronghang and Suris Coll-Vinent, Didac and Ryali, Chaitanya and Alwala, Kalyan Vasudev and Khedr, Haitham and Huang, Andrew and Lei, Jie and Ma, Tengyu and Guo, Baishan and Kalla, Arpit and Marks, Markus and Greer, Joseph and Wang, Meng and Sun, Peize and R{\"a}dle, Roman and Afouras, Triantafyllos and Mavroudi, Effrosyni and Xu, Katherine and Wu, Tsung-Han and Zhou, Yu and Momeni, Liliane and Hazra, Rishi and Ding, Shuangrui and Vaze, Sagar and Porcher, Francois and Li, Feng and Li, Siyuan and Kamath, Aishwarya and Cheng, Ho Kei and Doll{\'a}r, Piotr and Ravi, Nikhila and Saenko, Kate and Zhang, Pengchuan and Feichtenhofer, Christoph},
  booktitle = {International Conference on Learning Representations},
  title = {{SAM 3: Segment Anything with Concepts}},
  url = {https://proceedings.iclr.cc/paper_files/paper/2026/hash/e0982cbc81401df3430ee1ff780dc7a2-Abstract-Conference.html},
  year = {2026}
}

@inproceedings{lisa,
  author = {Lai, Xin and Tian, Zhuotao and Chen, Yukang and Li, Yanwei and Yuan, Yuhui and Liu, Shu and Jia, Jiaya},
  booktitle = {Proceedings of the IEEE/CVF Conference on Computer Vision and Pattern Recognition (CVPR)},
  pages = {9579--9589},
  title = {{LISA: Reasoning Segmentation via Large Language Model}},
  url = {https://openaccess.thecvf.com/content/CVPR2024/html/Lai_LISA_Reasoning_Segmentation_via_Large_Language_Model_CVPR_2024_paper.html},
  year = {2024}
}

@inproceedings{pixellm,
  author = {Ren, Zhongwei and Huang, Zhicheng and Wei, Yunchao and Zhao, Yao and Fu, Dongmei and Feng, Jiashi and Jin, Xiaojie},
  booktitle = {Proceedings of the IEEE/CVF Conference on Computer Vision and Pattern Recognition (CVPR)},
  pages = {26374--26383},
  title = {{PixelLM: Pixel Reasoning with Large Multimodal Model}},
  url = {https://openaccess.thecvf.com/content/CVPR2024/html/Ren_PixelLM_Pixel_Reasoning_with_Large_Multimodal_Model_CVPR_2024_paper.html},
  year = {2024}
}

@inproceedings{lora,
  author = {Hu, Edward J. and Shen, Yelong and Wallis, Phillip and Allen-Zhu, Zeyuan and Li, Yuanzhi and Wang, Shean and Wang, Lu and Chen, Weizhu},
  booktitle = {International Conference on Learning Representations},
  title = {{LoRA: Low-Rank Adaptation of Large Language Models}},
  url = {https://openreview.net/forum?id=nZeVKeeFYf9},
  year = {2022}
}

@misc{qwen35,
  author = {{Qwen Team}},
  howpublished = {Official model card},
  title = {{Qwen3.5-2B}},
  url = {https://huggingface.co/Qwen/Qwen3.5-2B},
  urldate = {2026-11-14},
  year = {2026}
}

@article{think2segrs,
  author = {Zhang, Xu and Ge, Junyao and Zheng, Yang and Guo, Kaitai and Liang, Jimin},
  doi = {10.1016/j.isprsjprs.2026.04.036},
  journal = {ISPRS Journal of Photogrammetry and Remote Sensing},
  pages = {217--235},
  title = {{Bridging semantics and geometry: A decoupled LVLM--SAM framework for reasoning segmentation in optical remote sensing}},
  url = {https://doi.org/10.1016/j.isprsjprs.2026.04.036},
  volume = {237},
  year = {2026}
}

@inproceedings{socioreasoner,
  author = {Wang, Yu and Wang, Yi and Dai, Rui and Wang, Yujie and Liu, Kaikui and Chu, Xiangxiang and Li, Yansheng},
  booktitle = {International Conference on Learning Representations},
  title = {{Urban Socio-Semantic Segmentation with Vision-Language Reasoning}},
  url = {https://proceedings.iclr.cc/paper_files/paper/2026/hash/792845ddfe4047d7066348e52e46b74d-Abstract-Conference.html},
  year = {2026}
}

@inproceedings{segagent,
  author = {Zhu, Muzhi and Tian, Yuzhuo and Chen, Hao and Zhou, Chunluan and Guo, Qingpei and Liu, Yang and Yang, Ming and Shen, Chunhua},
  booktitle = {Proceedings of the IEEE/CVF Conference on Computer Vision and Pattern Recognition (CVPR)},
  pages = {3686--3696},
  title = {{SegAgent: Exploring Pixel Understanding Capabilities in MLLMs by Imitating Human Annotator Trajectories}},
  url = {https://openaccess.thecvf.com/content/CVPR2025/html/Zhu_SegAgent_Exploring_Pixel_Understanding_Capabilities_in_MLLMs_by_Imitating_Human_CVPR_2025_paper.html},
  year = {2025}
}

@inproceedings{samveteran,
  author = {Du, Tianyuan and Li, Haopeng and Fan, Zhen and Zhang, Jiarui and Pan, Panwang and Zhang, Yang},
  booktitle = {International Conference on Learning Representations},
  title = {{SAM-Veteran: An MLLM-Based Human-like SAM Agent for Reasoning Segmentation}},
  url = {https://proceedings.iclr.cc/paper_files/paper/2026/hash/eb115a5de129c1b34c30dc5ee72c2ec7-Abstract-Conference.html},
  year = {2026}
}

@inproceedings{guideseg,
  author = {Vats, Vanshika and Rathee, Ashwani and Davis, James},
  booktitle = {Proceedings of the AAAI Conference on Artificial Intelligence},
  doi = {10.1609/aaai.v40i12.37923},
  number = {12},
  pages = {9612--9620},
  title = {{Guideline-Consistent Segmentation via Multi-Agent Refinement}},
  url = {https://doi.org/10.1609/aaai.v40i12.37923},
  volume = {40},
  year = {2026}
}

@inproceedings{cross,
  author = {Luo, Tingzhang and Liu, Ruizhong and Liu, Yichao and Fan, Cheng and Liu, Yu and Guo, Jianyuan},
  booktitle = {European Conference on Computer Vision},
  doi = {10.1007/978-3-032-37435-6_1},
  pages = {1--21},
  title = {{CROSS: Cascaded Distillation and Dual-Constraint Grounding for Remote Sensing Referring Segmentation}},
  url = {https://doi.org/10.1007/978-3-032-37435-6_1},
  year = {2026}
}

@inproceedings{rsagent,
  author = {He, Xingqi and Zhang, Yujie and Gao, Shuyong and Li, Wenjie and Hong, Lingyi and Chen, Mingxi and Jiang, Kaixun and Fu, Jiyuan and Zhang, Wenqiang},
  booktitle = {Proceedings of the 43rd International Conference on Machine Learning},
  pages = {42242--42266},
  publisher = {PMLR},
  title = {{{RSA}gent: Learning to Reason and Act via Multi-Turn Tool Invocations for Text-Guided Segmentation}},
  url = {https://proceedings.mlr.press/v306/he26av.html},
  volume = {306},
  year = {2026}
}

@article{rsrefseg2,
  author = {Chen, Keyan and Liu, Chenyang and Chen, Bowen and Zhang, Jiafan and Zou, Zhengxia and Shi, Zhenwei},
  doi = {10.1109/tgrs.2025.3647535},
  journal = {IEEE Transactions on Geoscience and Remote Sensing},
  pages = {1--20},
  title = {{RSRefSeg 2: Decoupling Referring Remote Sensing Image Segmentation With Foundation Models}},
  url = {https://doi.org/10.1109/tgrs.2025.3647535},
  volume = {64},
  year = {2026}
}

@inproceedings{focalclick,
  author = {Chen, Xi and Zhao, Zhiyan and Zhang, Yilei and Duan, Manni and Qi, Donglian and Zhao, Hengshuang},
  booktitle = {Proceedings of the IEEE/CVF Conference on Computer Vision and Pattern Recognition (CVPR)},
  pages = {1300--1309},
  title = {{FocalClick: Towards Practical Interactive Image Segmentation}},
  url = {https://openaccess.thecvf.com/content/CVPR2022/html/Chen_FocalClick_Towards_Practical_Interactive_Image_Segmentation_CVPR_2022_paper.html},
  year = {2022}
}

@article{iris,
  author = {Zhao, Yangang and Li, Jun and Wu, Bo and Zhao, Yehe and Mao, Zhihui},
  doi = {10.1109/lgrs.2025.3634831},
  journal = {IEEE Geoscience and Remote Sensing Letters},
  pages = {1--5},
  title = {{Iterative Refinement for Remote Sensing Image Segmentation via Instruction-Guided Closed-Loop Reasoning}},
  url = {https://doi.org/10.1109/lgrs.2025.3634831},
  volume = {23},
  year = {2026}
}

@article{segzero,
  author = {Liu, Yuqi and Peng, Bohao and Zhong, Zhisheng and Yue, Zihao and Lu, Fanbin and Yu, Bei and Jia, Jiaya},
  journal = {arXiv preprint arXiv:2503.06520},
  title = {{Seg-Zero: Reasoning-Chain Guided Segmentation via Cognitive Reinforcement}},
  url = {https://arxiv.org/abs/2503.06520},
  year = {2025}
}

@inproceedings{visionreasoner,
  author = {Liu, Yuqi and Qu, Tianyuan and Zhong, Zhisheng and Peng, Bohao and Liu, Shu and Yu, Bei and Jia, Jiaya},
  booktitle = {International Conference on Learning Representations},
  title = {{VisionReasoner: Unified Reasoning-Integrated Visual Perception via Reinforcement Learning}},
  url = {https://proceedings.iclr.cc/paper_files/paper/2026/hash/97c2f0fac182353062d304d0322ae285-Abstract-Conference.html},
  year = {2026}
}

@inproceedings{glamm,
  author = {Rasheed, Hanoona and Maaz, Muhammad and Shaji, Sahal and Shaker, Abdelrahman and Khan, Salman and Cholakkal, Hisham and Anwer, Rao M. and Xing, Eric and Yang, Ming-Hsuan and Khan, Fahad S.},
  booktitle = {Proceedings of the IEEE/CVF Conference on Computer Vision and Pattern Recognition (CVPR)},
  pages = {13009--13018},
  title = {{GLaMM: Pixel Grounding Large Multimodal Model}},
  url = {https://openaccess.thecvf.com/content/CVPR2024/html/Rasheed_GLaMM_Pixel_Grounding_Large_Multimodal_Model_CVPR_2024_paper.html},
  year = {2024}
}

@inproceedings{mmr,
  author = {Jang, Donggon and Cho, Yucheol and Lee, Suin and Kim, Taehyeon and Kim, Dae-Shik},
  booktitle = {International Conference on Learning Representations},
  title = {{MMR: A Large-scale Benchmark Dataset for Multi-target and Multi-granularity Reasoning Segmentation}},
  url = {https://proceedings.iclr.cc/paper_files/paper/2025/hash/a8661ab3440964c27794607f0ea73624-Abstract-Conference.html},
  year = {2025}
}

@inproceedings{remotereasoner,
  author = {Yao, Liang and Liu, Fan and Lu, Hongbo and Zhang, Chuanyi and Min, Rui and Xu, Shengxiang and Di, Shimin and Peng, Pai},
  booktitle = {Proceedings of the AAAI Conference on Artificial Intelligence},
  doi = {10.1609/aaai.v40i14.38175},
  number = {14},
  pages = {11883--11891},
  title = {{RemoteReasoner: Towards Unifying Geospatial Reasoning Workflow}},
  url = {https://doi.org/10.1609/aaai.v40i14.38175},
  volume = {40},
  year = {2026}
}

@article{segearthr1,
  author = {Li, Kaiyu and Xin, Zepeng and Pang, Li and Pang, Chao and Deng, Yupeng and Yao, Jing and Xia, Guisong and Meng, Deyu and Wang, Zhi and Cao, Xiangyong},
  journal = {arXiv preprint arXiv:2504.09644},
  title = {{SegEarth-R1: Geospatial Pixel Reasoning via Large Language Model}},
  url = {https://arxiv.org/abs/2504.09644},
  year = {2025}
}

@inproceedings{rs2sam2,
  author = {Rong, Fu and Lan, Meng and Zhang, Qian and Zhang, Lefei},
  booktitle = {Proceedings of the AAAI Conference on Artificial Intelligence},
  doi = {10.1609/aaai.v40i11.37828},
  number = {11},
  pages = {8751--8759},
  title = {{RS2-SAM2: Customized SAM2 for Referring Remote Sensing Image Segmentation}},
  url = {https://doi.org/10.1609/aaai.v40i11.37828},
  volume = {40},
  year = {2026}
}

@inproceedings{georeason,
  author = {Fan, Cheng and Wu, Han and Luo, Tingzhang and Dong, Minjing and Tang, Yehui and Guo, Jianyuan},
  booktitle = {Advances in Neural Information Processing Systems},
  note = {To appear},
  eprint = {2610.23416},
  archivePrefix = {arXiv},
  primaryClass = {cs.CV},
  title = {{GeoReason: Bridging Logical Reasoning and Spatial Fidelity in Remote Sensing Segmentation}},
  url = {https://neurips.cc/virtual/2026/poster/152062},
  year = {2026}
}

@inproceedings{sam2,
  author = {Ravi, Nikhila and Gabeur, Valentin and Hu, Yuan-Ting and Hu, Ronghang and Ryali, Chaitanya and Ma, Tengyu and Khedr, Haitham and R{\"a}dle, Roman and Rolland, Chloe and Gustafson, Laura and Mintun, Eric and Pan, Junting and Alwala, Kalyan Vasudev and Carion, Nicolas and Wu, Chao-Yuan and Girshick, Ross and Doll{\'a}r, Piotr and Feichtenhofer, Christoph},
  booktitle = {International Conference on Learning Representations},
  title = {{SAM 2: Segment Anything in Images and Videos}},
  url = {https://proceedings.iclr.cc/paper_files/paper/2025/hash/45c1f6a8cbf2da59ebf2c802b4f742cd-Abstract-Conference.html},
  year = {2025}
}

@article{refsegrs,
  author = {Yuan, Zhenghang and Mou, Lichao and Hua, Yuansheng and Zhu, Xiao Xiang},
  doi = {10.1109/tgrs.2024.3369720},
  journal = {IEEE Transactions on Geoscience and Remote Sensing},
  pages = {1--12},
  title = {{RRSIS: Referring Remote Sensing Image Segmentation}},
  url = {https://doi.org/10.1109/tgrs.2024.3369720},
  volume = {62},
  year = {2024}
}

@article{crobim,
  author = {Dong, Zhe and Sun, Yuzhe and Liu, Tianzhu and Zuo, Wangmeng and Gu, Yanfeng},
  journal = {arXiv preprint arXiv:2410.08613},
  title = {{Cross-Modal Bidirectional Interaction Model for Referring Remote Sensing Image Segmentation}},
  url = {https://arxiv.org/abs/2410.08613},
  year = {2024}
}

@inproceedings{attention,
  author = {Vaswani, Ashish and Shazeer, Noam and Parmar, Niki and Uszkoreit, Jakob and Jones, Llion and Gomez, Aidan N. and Kaiser, {\L}ukasz and Polosukhin, Illia},
  booktitle = {Advances in Neural Information Processing Systems},
  title = {{Attention is All you Need}},
  url = {https://papers.nips.cc/paper/7181-attention-is-all-you-need},
  volume = {30},
  year = {2017}
}

@inproceedings{lovasz,
  author = {Berman, Maxim and Triki, Amal Rannen and Blaschko, Matthew B.},
  booktitle = {Proceedings of the IEEE Conference on Computer Vision and Pattern Recognition (CVPR)},
  pages = {4413--4421},
  title = {{The Lov{\'a}sz-Softmax Loss: A Tractable Surrogate for the Optimization of the Intersection-Over-Union Measure in Neural Networks}},
  url = {https://openaccess.thecvf.com/content_cvpr_2018/html/Berman_The_LovaSz-Softmax_Loss_CVPR_2018_paper.html},
  year = {2018}
}

@inproceedings{dagger,
  author = {Ross, Stephane and Gordon, Geoffrey and Bagnell, Drew},
  booktitle = {Proceedings of the Fourteenth International Conference on Artificial Intelligence and Statistics},
  pages = {627--635},
  publisher = {PMLR},
  title = {{A Reduction of Imitation Learning and Structured Prediction to No-Regret Online Learning}},
  url = {https://proceedings.mlr.press/v15/ross11a.html},
  volume = {15},
  year = {2011}
}

@article{dinkelbach,
  author = {Dinkelbach, Werner},
  doi = {10.1287/mnsc.13.7.492},
  journal = {Management Science},
  number = {7},
  pages = {492--498},
  title = {{On Nonlinear Fractional Programming}},
  url = {https://doi.org/10.1287/mnsc.13.7.492},
  volume = {13},
  year = {1967}
}

@inproceedings{positionbias,
  author = {Li, Zongjie  and
Wang, Chaozheng  and
Ma, Pingchuan  and
Wu, Daoyuan  and
Wang, Shuai  and
Gao, Cuiyun  and
Liu, Yang},
  booktitle = {Proceedings of the 2024 Conference on Empirical Methods in Natural Language Processing},
  doi = {10.18653/v1/2024.emnlp-main.621},
  pages = {11084--11108},
  publisher = {Association for Computational Linguistics},
  title = {{Split and Merge: Aligning Position Biases in {LLM}-based Evaluators}},
  url = {https://aclanthology.org/2024.emnlp-main.621/},
  year = {2024}
}

@inproceedings{gkd,
  author = {Agarwal, Rishabh and Vieillard, Nino and Zhou, Yongchao and Stanczyk, Piotr and Ramos Garea, Sabela and Geist, Matthieu and Bachem, Olivier},
  booktitle = {International Conference on Learning Representations},
  title = {{On-Policy Distillation of Language Models: Learning from Self-Generated Mistakes}},
  url = {https://proceedings.iclr.cc/paper_files/paper/2024/hash/5be69a584901a26c521c2b51e40a4c20-Abstract-Conference.html},
  year = {2024}
}

@inproceedings{ste,
  author = {Yin, Penghang and Lyu, Jiancheng and Zhang, Shuai and Osher, Stanley and Qi, Yingyong and Xin, Jack},
  booktitle = {International Conference on Learning Representations},
  title = {{Understanding Straight-Through Estimator in Training Activation Quantized Neural Nets}},
  url = {https://openreview.net/forum?id=Skh4jRcKQ},
  year = {2019}
}

@inproceedings{adamw,
  author = {Loshchilov, Ilya and Hutter, Frank},
  booktitle = {International Conference on Learning Representations},
  title = {{Decoupled Weight Decay Regularization}},
  url = {https://openreview.net/forum?id=Bkg6RiCqY7},
  year = {2019}
}

@inproceedings{sam,
  author = {Kirillov, Alexander and Mintun, Eric and Ravi, Nikhila and Mao, Hanzi and Rolland, Chloe and Gustafson, Laura and Xiao, Tete and Whitehead, Spencer and Berg, Alexander C. and Lo, Wan-Yen and Dollar, Piotr and Girshick, Ross},
  booktitle = {Proceedings of the IEEE/CVF International Conference on Computer Vision (ICCV)},
  pages = {4015--4026},
  title = {{Segment Anything}},
  url = {https://openaccess.thecvf.com/content/ICCV2023/html/Kirillov_Segment_Anything_ICCV_2023_paper.html},
  year = {2023}
}

@inproceedings{geochat,
  author = {Kuckreja, Kartik and Danish, Muhammad Sohail and Naseer, Muzammal and Das, Abhijit and Khan, Salman and Khan, Fahad Shahbaz},
  booktitle = {Proceedings of the IEEE/CVF Conference on Computer Vision and Pattern Recognition (CVPR)},
  pages = {27831--27840},
  title = {{GeoChat: Grounded Large Vision-Language Model for Remote Sensing}},
  url = {https://openaccess.thecvf.com/content/CVPR2024/html/Kuckreja_GeoChat_Grounded_Large_Vision-Language_Model_for_Remote_Sensing_CVPR_2024_paper.html},
  year = {2024}
}

@inproceedings{vrsbench,
  author = {Li, Xiang and Ding, Jian and Elhoseiny, Mohamed},
  booktitle = {Advances in Neural Information Processing Systems},
  doi = {10.52202/079017-0106},
  pages = {3229--3242},
  title = {{VRSBench: A Versatile Vision-Language Benchmark Dataset for Remote Sensing Image Understanding}},
  url = {https://proceedings.neurips.cc/paper_files/paper/2024/hash/05b7f821234f66b78f99e7803fffa78a-Abstract-Datasets_and_Benchmarks_Track.html},
  volume = {37},
  year = {2024}
}

@article{sa2va,
  author = {Yuan, Haobo and Li, Xiangtai and Zhang, Tao and Sun, Yueyi and Huang, Zilong and Xu, Shilin and Ji, Shunping and Tong, Yunhai and Qi, Lu and Feng, Jiashi and Yang, Ming-Hsuan},
  doi = {10.1109/tpami.2026.3720379},
  journal = {IEEE Transactions on Pattern Analysis and Machine Intelligence},
  number = {11},
  pages = {14922--14936},
  title = {{Sa2VA: Marrying SAM2 With MLLM for Dense Grounded Understanding of Images and Videos}},
  url = {https://doi.org/10.1109/tpami.2026.3720379},
  volume = {48},
  year = {2026}
}

\clearpage
\onecolumn
\appendix
\setcounter{figure}{0}
\setcounter{table}{0}
\setcounter{equation}{0}
\setcounter{algorithm}{0}
\setcounter{proposition}{0}
\renewcommand{\theHfigure}{supp.\arabic{figure}}
\renewcommand{\theHtable}{supp.\arabic{table}}
\renewcommand{\theHequation}{supp.\arabic{equation}}
\providecommand{\theHalgorithm}{supp.\arabic{algorithm}}
\setcounter{tocdepth}{2}
\addtocontents{toc}{\protect\setcounter{tocdepth}{2}}
\setcounter{totalnumber}{12}
\setcounter{topnumber}{8}
\setcounter{bottomnumber}{8}
\renewcommand{\floatpagefraction}{0.65}
\begin{center}
{\Large\bfseries GeoReVis: Supplementary Material\par}
\end{center}
\vspace{1em}
This supplement follows the argument of the main paper. Section~\ref{sec:supp_insight} develops the evidence for Revision Degradation and Evidence Fragmentation. Section~\ref{sec:supp_arch} gives the complete method and its mathematical foundations. Sections~\ref{sec:supp_train} and~\ref{sec:supp_eval} specify supervision, training, and evaluation. Section~\ref{sec:supp_results} provides extended results and visual analyses.

\begingroup
\hypersetup{linkcolor=black}
\tableofcontents
\endgroup
\twocolumn
\section{Evidence Behind the Two Insights}
\label{sec:supp_insight}
\subsection{Revision Degradation Despite Local Repair}
A revision changes both the target pixels recovered and the errors introduced elsewhere. To separate these effects, let $M$ and $H$ denote the initial and revised masks, and let $G$ be the target. The corrected pixel set is
\begin{equation}
\mathcal F=((H\setminus M)\cap G)\cup((M\setminus H)\setminus G).
\label{eq:supp_fixed_pixels}
\end{equation}
The newly erroneous pixel set is
\begin{equation}
\mathcal D=((M\setminus H)\cap G)\cup((H\setminus M)\setminus G).
\label{eq:supp_fix_damage}
\end{equation}
The two terms in $\mathcal F$ recover missing target pixels and remove background. Those in $\mathcal D$ remove correct target pixels and add background. Thus, repairing a visible error and improving the complete mask are different outcomes.

Table~\ref{tab:supp_insight} extends the main paper's Fig.~1. GeoReVis-R is GeoReVis with revision only. SocioReasoner$^\ast$ uses Qwen3.5-2B and frozen SAM3 trained on LaSeRS, with the optical image and instruction as input. Both are evaluated on all 1,158 LaSeRS samples.

Revision Degradation, defined by an IoU decrease exceeding 0.01, occurs in 27.81\% and 19.34\% of samples. Their positive mean gains therefore coexist with substantial per-sample damage. This is the empirical basis for Candidate Selection: the executed mask, rather than the proposed correction alone, determines whether an update is useful.
\begin{table}[!t]\centering
\caption{\textbf{Evidence for the two insights.} Scores and rates are in \%. Macro scores follow main Fig.~1. Query statistics use 1,158 samples, and capacity statistics use 1,565 GT-aligned targets. ``Gain with target loss'' is conditional on IoU gain $>0.01$. $\mathcal F$ and $\mathcal D$ are defined in Eqs.~\ref{eq:supp_fixed_pixels} and~\ref{eq:supp_fix_damage}. The last block evaluates alternative compositions within each target's candidate pool.}\label{tab:supp_insight}
\footnotesize\setlength{\tabcolsep}{2.5pt}\renewcommand{\arraystretch}{1.08}
\begin{tabularx}{\linewidth}{>{\raggedright\arraybackslash}Xcc}\toprule
\tablehead{Quantity} & \tablehead{GeoReVis-R} & \tablehead{SocioReasoner$^\ast$}\\\midrule
\multicolumn{3}{c}{\textit{LaSeRS macro gIoU: main Fig.~1}}\\\midrule
Initial Segmentation & 63.28 & 62.26\\
Revised & 64.64 & 63.13\\
Optimal Candidate & 66.94 & 63.90\\
Regional Oracle & 75.39 & 68.97\\
\rowcolor{paperteal!9}With Full GeoReVis & 72.00 & 66.45\\\midrule
\multicolumn{3}{c}{\textit{Revision behavior: query-level statistics}}\\\midrule
Mean IoU, initial $\rightarrow$ revised & 65.34 $\rightarrow$ 65.61 & 57.12 $\rightarrow$ 57.33\\
IoU gain $>0.01$ & 38.69 & 24.87\\
IoU change within $\pm0.01$ & 33.51 & 55.79\\
IoU loss $>0.01$ & 27.81 & 19.34\\
Near-stable: both $|\mathcal F|,|\mathcal D|>0.01|G|$ & 31.44 & 26.78\\
Both repair and damage & 88.95 & 85.41\\
Gain with target loss $>0.01|G|$ & 34.15 & 21.53\\
Optimal Candidate omits $>0.01|G|$ & 35.15 & 22.97\\\midrule
\multicolumn{3}{c}{\textit{Compositional capacity: target-level statistics}}\\\midrule
Mean Optimal Candidate IoU & 68.36 & 60.12\\
Oracle Subset Union gain & +0.55 & +0.31\\
Regional Oracle gain & +4.97 & +3.18\\
Targets with regional gain $>0.01$ & 45.11 & 33.67\\
Targets with subset-union gain $>0.01$ & 6.33 & 4.09\\
\bottomrule\end{tabularx}\end{table}

A near-stable revision has an absolute IoU change of at most 0.01. Among these samples, 31.44\% for GeoReVis-R and 26.78\% for SocioReasoner$^\ast$ have both $|\mathcal F|$ and $|\mathcal D|$ exceeding 1\% of $|G|$. A small net change can therefore conceal meaningful repair and damage. Mask feedback needs to be followed by an explicit comparison of what the execution changed.

\subsection{Evidence Fragmentation Beyond Candidate Selection}
Let $\mathcal H$ contain all observed candidates. For offline analysis, the optimal candidate is
\begin{equation}
H^\star=\arg\max_{H\in\mathcal H}\IoU(H,G).
\end{equation}
The target evidence available outside this mask is
\begin{equation}
E^\star=\left(G\cap\bigcup_{H\in\mathcal H}H\right)\setminus H^\star.
\label{eq:supp_missing_evidence}
\end{equation}
A sample exhibits measurable omission when $|E^\star|>0.01|G|$. Such omissions occur in 35.15\% and 22.97\% of the evaluated samples. The evidence already exists in the candidate bank, so generating a better whole mask is not the only route to improvement.

The regional oracle makes the distinction concrete. Under the LaSeRS macro aggregation used in the main figure, GeoReVis-R's optimal candidate reaches 66.94\% gIoU, while regional composition reaches 75.39\%. For SocioReasoner$^\ast$, these scores are 63.90\% and 68.97\%. The gaps are 8.45\% and 5.07\%, respectively. They quantify the quality available from recombining observed regions beyond whole-mask selection. On 1,565 GT-aligned targets, regional recombination improves on the optimal candidate more often than a union of complete candidates. This supports the use of Membership Regions as the decision unit.

\paragraph{The river example isolates the role of local evidence.}
Main Fig.~1(b) illustrates river segmentation near a bridge on LaSeRS image \texttt{04393.jpg}. The 83.1\%-IoU candidate captures more of the requested river overall, yet omits a bridge-adjacent segment retained by the 66.9\%-IoU candidate. Their ordering is by whole-mask quality, not chronological revision. The example illustrates why an optimal candidate can still discard useful evidence. Its illustrative masks on real imagery are distinguished from the aggregate model statistics and the actual prediction examples in Sec.~\ref{sec:supp_recovery_case}.

\paragraph{Aggregation follows the question being evaluated.}
The LaSeRS macro scores average its nine evaluation dimensions. Revision frequencies and mean query IoUs use 1,158 queries. Capacity analyses use 1,565 GT-aligned targets. Table~\ref{tab:supp_insight} separates these units, so each statistic measures either observed behavior or the compositional capacity of its stated candidate pool.

\section{Complete Method and Mathematical Foundations}
\label{sec:supp_arch}
\label{sec:supp_props}
\subsection{Target Decomposition and Initial Segmentation}
Given image $I$ and complete instruction $q$, Target Decomposition produces target descriptions $q_1,\ldots,q_L$. Each description retains the relevant entity, relation, part, and collection constraints. Spatial Grounding converts $q_l$ into boxes and signed points, and frozen SAM3~\cite{sam3} produces the corresponding initial mask. The full instruction remains visible to later roles, so a part-level edit is interpreted within the requested scene-level relationship.

All learned roles use Qwen3.5-2B~\cite{qwen35} with role-specific LoRA adapters~\cite{lora}. The grounding adapter is merged into the revision backbone. Candidate Selection is trained on the original Qwen backbone and then adapted to multi-round observations. Cross-Mask Evidence Consolidation initializes from this multi-round selector. Feedback Assessment has a separate adapter. This allocation lets the system learn distinct decisions while sharing the same backbone architecture.

For clarity, the following describes one target. Its state at round $t$ is $(M_{t-1},\mathcal H_{t-1},h_{t-1})$, where $M_{t-1}$ is the committed mask, $\mathcal H_{t-1}$ stores observed masks, and $h_{t-1}$ summarizes earlier feedback and decisions. A Concept Candidate is generated once from a semantic prompt and cached. Spatial grounding establishes the initial answer, while the concept candidate offers an alternative extent for connected targets.

\subsection{Selective Feedback Revision}
\paragraph{Executable actions localize the intended change.}
Feedback Assessment examines the image, full instruction, current mask, and decision history. It identifies missing regions, excess regions, and contextual mismatches. The reviser maps this observation to an action $a=(o,\mathcal B,\mathcal P)$, consisting of an operation, editing boxes, and signed points. SAM3 returns support $E_a$. With $D_a$ denoting the union of editing boxes, the executed candidate is
\begin{equation}
H_a=\begin{cases}
E_a,&o=\mathrm{replace},\\
M_{t-1}\cup(E_a\cap D_a),&o=\mathrm{add},\\
M_{t-1}\setminus(E_a\cap D_a),&o=\mathrm{erase}.
\end{cases}
\label{eq:supp_execute}
\end{equation}
For addition and erasure, the changed pixels satisfy
\begin{equation}
H_a\triangle M_{t-1}\subseteq D_a,
\end{equation}
where $\triangle$ denotes symmetric difference. Pixels outside the designated editing extent are unchanged. The operation type and spatial extent therefore provide direct control over local correction.

\paragraph{Candidate Selection evaluates actual pixel changes.}
The candidate pool $\mathcal P_t$ contains the committed mask and executed revisions, with the Concept Candidate added in the first round. The selector observes full-mask overlays and aligned difference views under the complete instruction. Global views determine target identity and relations. Difference views reveal added and removed support.

\paragraph{An ordered comparison pass selects the reference.}
Let $(H_{t,1},\ldots,H_{t,J_t})$ be the ordered challengers. Starting from $B_t^{(0)}=M_{t-1}$, Candidate Selection updates
\begin{equation}
B_t^{(j)}=\operatorname{PairSelect}_{\psi}
 (I,q,B_t^{(j-1)},H_{t,j}).
\label{eq:supp_ordered_selection}
\end{equation}
The final incumbent $B_t^{(J_t)}$ becomes the reference $B_t$.
The first round compares the initial mask with its executed revision, then compares the winner with the Concept Candidate. Later rounds process the learned revision first, followed by available local addition and erasure candidates. The local actions use the largest observed outside-mask and inside-mask regions, respectively. Repeated state--action pairs and duplicate masks are skipped. The cached Concept Candidate remains available for feedback and evidence consolidation.

Each pair is assigned A/B labels by a fixed pseudorandom mapping keyed by the query, round, and comparison index. The larger preference logit determines the winner, with A resolving an exact logit tie. Traversal order is fixed before comparison and uses no GT scores.

Preferences may form a cycle, for example,
\begin{equation}
H_1\succ H_2\succ H_3\succ H_1.
\end{equation}
The final incumbent after the single pass is the reference. This rule gives a definite outcome in at most $J_t$ comparisons and keeps selection cost linear in the number of challengers.

Every executed candidate enters the History Bank, including candidates not selected as $B_t$. Whole-mask quality decides the reference, while observed geometry determines the evidence available for consolidation. These decisions serve different purposes: selecting a reliable current answer and preserving local evidence for future use.

\paragraph{A single committed state closes the feedback loop.}
Algorithm~\ref{alg:georevis} connects the complete procedure. The adopted mask becomes the next round's observation. All alternatives remain in the same evidence bank. Three rounds are performed for each target, and the final target masks are merged into the requested segmentation.
\begin{algorithm}[t]
\caption{GeoReVis inference for one target}
\label{alg:georevis}
\begin{algorithmic}[1]
\Require Image $I$, complete instruction $q$, target description $q_l$
\State $M\gets\operatorname{InitialSegmentation}(I,q_l)$
\State $H_c\gets\operatorname{ConceptCandidate}(I,q_l)$
\State $\mathcal H\gets\{M,H_c\}$, $h\gets\emptyset$
\For{$t=1,2,3$}
\State Obtain feedback from $(I,q,M,h)$
\State Generate actions and execute them to obtain $\mathcal R_t$
\State $\mathcal L_t\gets$ available revisions in learned/add/erase order
\State Append $H_c$ to $\mathcal L_t$ only if $t=1$
\State $\mathcal P_t\gets\{M\}\cup\mathcal L_t$, $B\gets M$
\For{each challenger $H$ in $\mathcal L_t$}
\State $B\gets\operatorname{PairSelect}_\psi(I,q,B,H)$ if $H\ne B$
\EndFor
\State $\mathcal H\gets\operatorname{Deduplicate}(\mathcal H\cup\mathcal P_t)$
\State Partition $\mathcal H$ into $C,R_1,\ldots,R_K$
\State Encode shared views and jointly predict inclusion code $x$
\State $S\gets C+\sum_{i=1}^{K}x_iR_i$
\State Compare $(S,B)$ and $(B,S)$ to obtain margin $m$
\State $M\gets S$ if $m>0$, otherwise $M\gets B$
\State $\mathcal H\gets\mathcal H\cup\{S\}$ and update history $h$
\EndFor
\State \Return $M$
\end{algorithmic}
\end{algorithm}

\subsection{Membership Regions Preserve Candidate Evidence}
\label{sec:supp_representation}
For the deduplicated bank $\mathcal H=\{H_1,\ldots,H_J\}$, each pixel has the membership signature
\begin{equation}
s(p)=[H_1(p),\ldots,H_J(p)].
\end{equation}
The common foreground and observed support are, respectively,
\begin{align}
C&=\bigcap_jH_j,\\
U&=\bigcup_jH_j.
\end{align}
Within $U\setminus C$, pixels with the same signature are partitioned into eight-connected regions $R_1,\ldots,R_K$. No minimum-area filtering removes small regions. This preserves narrow structures and small parts that are important in remote sensing.

We use the same symbol for a foreground pixel set and its binary indicator mask. Written in binary-mask form, the compositional family is
\begin{equation}
\mathcal S(\mathcal H)=\left\{C+\sum_{i=1}^{K}x_iR_i:x\in\{0,1\}^K\right\}.
\label{eq:supp_family}
\end{equation}
Each inclusion bit $x_i$ retains or drops $R_i$. The regions and $C$ are pairwise disjoint, so their pixelwise sum remains binary and equals the union of the retained regions and common foreground. All reconstructed masks satisfy $C\subseteq S\subseteq U$. Revision supplies new geometry and changes membership patterns. Consolidation then makes use of the complementary evidence within that observed space.

\begin{proposition}[Candidate representability]
Every source mask in $\mathcal H$, including the selected reference $B$, belongs to $\mathcal S(\mathcal H)$.
\end{proposition}
\begin{proof}
A source mask contains all of $C$ and no pixel outside $U$. Its membership is constant on every $R_i$. Assigning $x_i=H_j(p)$ for any $p\in R_i$ therefore reconstructs $H_j$ exactly.
\end{proof}
This property preserves whole-mask selection as a feasible composition while admitting new combinations of its parts.

\begin{proposition}[History expansion]
Expanding the candidate bank preserves the compositional family:
\begin{equation}
\mathcal H\subseteq\mathcal H'
\quad\Longrightarrow\quad
\mathcal S(\mathcal H)\subseteq\mathcal S(\mathcal H').
\end{equation}
\end{proposition}
\begin{proof}
Adding a candidate extends each membership signature, so every old disagreement region splits into connected children or remains unchanged. Assign its inclusion bit to each child. Any part of the old common foreground that becomes variable is retained by selecting its children. Newly observed support is excluded. This reconstructs every old feasible mask using the new partition.
\end{proof}
Consequently, the best attainable IoU within this family is nondecreasing as evidence accumulates. This representational property explains why rejected candidates remain in the bank: later decisions can reuse their regions without losing earlier feasible answers.

\subsection{Shared Encoding and Joint Region Prediction}
\paragraph{Three views provide context at complementary scales.}
The Shared Encoder receives the complete image, a reference overlay cropped around the candidate union, and a zoomed reference overlay. Crop coordinates come from observed masks. For view $v$, let $V_v$ be its image extent and $F_v(p)$ its visual feature field. Region pooling gives
\begin{equation}
r_i^{(v)}=\frac{1}{|R_i\cap V_v|}\sum_{p\in R_i\cap V_v}F_v(p).
\label{eq:supp_pool}
\end{equation}
The corresponding coverage is
\begin{equation}
c_i^{(v)}=\frac{|R_i\cap V_v|}{|R_i|}.
\end{equation}
An empty intersection contributes a zero feature and zero coverage. Pooling follows the native pixel-to-token mapping, with patch contributions weighted by their overlap with the region. The coverage values distinguish absent evidence in a crop from a low visual response.

We also compute a bounded region-to-reference size ratio:
\begin{equation}
\rho_i=\frac{|R_i|}{|R_i|+|B|}.
\end{equation}
Each region concatenates its three pooled features and coverage values with normalized center and box coordinates, area relative to $U$, the ratio $\rho_i$, an $8\times8$ occupancy descriptor, and its reference inclusion bit $b_i$.

The Shared Encoder also retains all valid image and text hidden states in context $Z$, prefixed by a full-context summary $z_q$.

\paragraph{Joint prediction links each region to the complete composition.}
A width-256 projection forms each region query:
\begin{equation}
h_i=\operatorname{LN}(W_r[r_i,g_i,b_i]+W_qz_q).
\end{equation}
Here $r_i$ collects pooled features and coverage, and $g_i$ is the geometry and occupancy descriptor. Four-head cross-attention~\cite{attention} connects the query to the full image and instruction context $Z$:
\begin{equation}
z_i=h_i+W_o\operatorname{MHA}(h_i,Z,Z).
\end{equation}
An area-weighted summary aggregates evidence across regions:
\begin{equation}
\bar z=\frac{\sum_i|R_i|z_i}{\sum_i|R_i|}.
\end{equation}
The readout combines each region with this summary:
\begin{equation}
f(e)_i=\operatorname{MLP}([z_i,\bar z]).
\label{eq:supp_joint_readout}
\end{equation}
Here $e$ denotes the shared inputs. The complete instruction determines semantic relevance, while the summary lets each decision account for the spatial impact of other regions.

All regions share one multimodal encoding, avoiding a separate VLM query for each fragment. Attention is processed in chunks of 256 region queries with unchanged shared keys and values.

\paragraph{Reference anchoring preserves the starting prediction.}
The live readout $f_\phi$ and its frozen initial copy $f_{\bar\phi}$ receive the same current encoder features. Inclusion logits are
\begin{equation}
\ell_i=2b_i-1+f_\phi(e)_i-f_{\bar\phi}(e)_i.
\label{eq:supp_anchor}
\end{equation}
At initialization, $\phi_0=\bar\phi$, so the two outputs cancel for every input and thresholding reproduces $B$ exactly. Zero logits retain the corresponding reference bits. Only the anchor's parameters are frozen. Its feature path remains differentiable, giving
\begin{equation}
\frac{\partial\ell}{\partial\theta}
=\left(\frac{\partial f_\phi}{\partial e}-\frac{\partial f_{\bar\phi}}{\partial e}\right)
\frac{\partial e}{\partial\theta}.
\label{eq:supp_anchor_gradient}
\end{equation}
Here $\theta$ denotes the consolidation adapter. The regional encoder gradients initially cancel, while the live readout can learn immediately. As the readout changes, regional supervision also adapts the shared representation. Relative Adoption supplies its own learning signal to the adapter from the start.

\subsection{Regional Decisions and Complete-Mask Quality}
\label{sec:supp_netgain}
Let the nonempty GT have area $T=|G|$. The target areas in the common foreground and each region are
\begin{align}
\tau_0&=|C\cap G|,\\
\tau_i&=|R_i\cap G|.
\end{align}
Their background areas are
\begin{align}
\varphi_0&=|C\setminus G|,\\
\varphi_i&=|R_i\setminus G|.
\end{align}
The IoU of code $x$ on this pixel grid is
\begin{equation}
Q(x)=\frac{N(x)}{D(x)}
=\frac{\tau_0+\sum_i\tau_ix_i}{T+\varphi_0+\sum_i\varphi_ix_i}.
\label{eq:supp_fraction}
\end{equation}
For any two codes $x,x'$, direct subtraction gives the exact identity
\begin{equation}
Q(x')-Q(x)=
\frac{\sum_i(x_i'-x_i)[\tau_i-Q(x)\varphi_i]}{D(x')}.
\label{eq:supp_netgain}
\end{equation}
The contribution of a region depends on its target pixels, background pixels, and current complete-mask quality. Several local changes can reinforce or offset one another. The identity therefore motivates supervising the complete reconstruction in addition to regional semantics.

For a continuous inclusion $p_i\in[0,1]$, the corresponding derivative is
\begin{equation}
\frac{\partial Q(p)}{\partial p_i}=\frac{\tau_i-Q(p)\varphi_i}{D(p)}.
\label{eq:supp_gradient}
\end{equation}
It rewards the same target-versus-background tradeoff as the discrete gain identity. This provides a direct connection between joint mask-quality supervision and local inclusion decisions.

\paragraph{An exact regional optimum supplies training labels.}
At optimal ratio $Q^*$, fractional optimization~\cite{dinkelbach} gives
\begin{equation}
\max_x[N(x)-Q^*D(x)]=0.
\label{eq:supp_fractional_optimality}
\end{equation}
The expression is separable in $x_i$. A region has positive contribution exactly when
\begin{equation}
\tau_i-Q^*\varphi_i>0.
\end{equation}
Equivalently, its target purity satisfies
\begin{equation}
\frac{\tau_i}{\tau_i+\varphi_i}>\frac{Q^*}{1+Q^*}.
\end{equation}
Such regions are included, while those with negative contribution are excluded. Sorting regions by purity and evaluating all prefixes, including the empty prefix, finds an exact optimum in $O(K\log K)$ time. Regions with zero contribution are excluded to minimize background among equal-IoU solutions.

The result is computed on the native grid for regional supervision. The principal training objective separately scores the full mask on the benchmark evaluation grid.

\subsection{Relative Adoption and Closed-Loop Improvement}
The consolidation adapter compares reconstruction $S$ with reference $B$ in both input orders. Let $m^{SB}$ and $m^{BS}$ always denote the reconstruction logit minus the reference logit in their respective presentations. The averaged margin is
\begin{equation}
m=\frac{m^{SB}+m^{BS}}{2}.
\end{equation}
It selects $S$ when positive and $B$ otherwise. Reversing the presentation removes reliance on one candidate position~\cite{positionbias}. Whether adopted or not, $S$ enters the History Bank.

For a non-neutral pair, define its IoU difference as
\begin{equation}
\Delta=\IoU(S,G)-\IoU(B,G).
\end{equation}
Its preferred direction is $y=\operatorname{sign}(\Delta)$.
The realized quality satisfies
\begin{equation}
\IoU(M,G)=\max\{\IoU(S,G),\IoU(B,G)\}
-|\Delta|\mathbf1[\mathrm{wrong}].
\label{eq:supp_adoption_identity}
\end{equation}
Thus, improving the reconstruction and making the correct adoption are complementary requirements. The quality cost of a wrong decision is exactly $|\Delta|$. A wrong decision implies $ym\leq0$, so
\begin{equation}
|\Delta|\mathbf1[\mathrm{wrong}]
\leq\frac{|\Delta|}{\log2}\log(1+e^{-ym}).
\label{eq:supp_adoption_bound}
\end{equation}
Weighting preference learning by $|\Delta|$ therefore connects its surrogate to the segmentation quality at stake. Convexity also gives
\begin{equation}
\log(1+e^{-ym})\leq\tfrac12\sum_{o\in\{SB,BS\}}\log(1+e^{-ym^o}).
\end{equation}
Training both orders controls the same averaged margin used at inference. The empirical Degradation Rate and Correct-Pixel Preservation then assess the learned decisions over complete trajectories.

\section{Execution-Driven Data Pipeline and Learning}
\label{sec:supp_train}
\subsection{Supervision Follows Actual Execution}
The pipeline labels the decisions that GeoReVis actually makes: which correction to execute, which candidate to select, which regions to combine, and whether to adopt the reconstruction. It first constructs each prediction through the same interfaces used at inference. Training GT then evaluates the resulting masks. This makes supervision specific to both the model's observed states and the available actions.

\paragraph{Initial Segmentation establishes the target interface.}
Target Decomposition learns from 29,201 instruction records, while Spatial Grounding uses 38,320 demonstrations. These roles produce target descriptions and spatial prompts. Their outputs establish the starting masks for subsequent execution-driven supervision.

\paragraph{Executed actions teach effective Feedback Revision.}
Teacher proposals, proposal-seeded search, and rules generate spatial edits. Every proposal is executed through SAM3, with utility
\begin{equation}
\Delta_a=\IoU(H_a,G)-\IoU(M,G).
\end{equation}
Programs with $\Delta_a\geq0.01$ and valid signed prompts supervise the reviser. Verification connects the proposed operation to its actual segmentation effect.

The verified pool contains 32,435 programs: 2,297 direct external proposals, 22,056 proposal-seeded search outcomes, and 8,082 rule-generated corrections. MiniMax-M3.1 supplies the external proposals during training.

\paragraph{Proposal-seeded search refines executable spatial hypotheses.}
Each source state contains the image, complete target context, current predicted mask, and training annotation. The teacher proposes up to 18 programs using \emph{replace}, \emph{add}, or \emph{erase}. A program contains at most four independent boxes, four positive points, and two negative points in total, with at least one positive point per box. Coordinates are normalized to $[0,1000]$. Every program acts on the same original state, allowing its realized gain to be compared directly.

After execution, the pool combines teacher proposals with available rule and geometric-scope candidates. Programs are deduplicated and ranked by executed IoU. The three highest-ranked programs with valid signed prompts seed one local search pass. For each seed, the ordered variants are:
\begin{enumerate}[leftmargin=*,nosep]
\item \textbf{Box extent.} Contract each box edge by 0.5 image pixels, expand it by 1 pixel, or scale its width and height about the center by 0.95 or 1.05. Boxes are clipped to the image and adjusted to retain their positive points.
\item \textbf{Editing scope.} For an addition or replacement seed, swap \emph{add} and \emph{replace} while retaining its spatial prompts.
\item \textbf{Background exclusion.} For these seeds, identify up to two false-positive components in the executed mask. Each yields a separate variant with an interior negative point in the smallest seed box containing that point, subject to the two-negative-point budget.
\end{enumerate}
Invalid or duplicate variants are removed. The first 18 unique variants are executed, and the local pass ends when this budget or the finite variant list is exhausted. The 22,056 proposal-seeded labels are those whose seed is an external teacher proposal. This procedure converts semantic proposals into spatial supervision verified by the actual segmentation decoder.

\paragraph{Bounded supplementation covers residual errors.}
Geometric-scope proposals group five to eight target components into two to four prompt regions, favoring balanced groups and tight boxes, with at most 18 programs. After local search, one residual pass tests at most 18 additional programs that jointly add missing components, erase excess components, or group larger target collections. This pass is skipped when the best valid candidate or initial mask already reaches 0.95 IoU. Search therefore ends after fixed candidate lists rather than recursively refining new variants.

Candidate retention requires a gain of at least 0.01 and operation-consistent point signs. For addition and replacement, positive points lie on the target and negative points on background. For erasure, positive points identify unwanted current support and negative points lie outside it.

Up to three distinct near-best alternatives per state are retained. Let $Q_0$ and $Q_{\max}$ be the initial and best verified IoUs. Each retained candidate must satisfy both thresholds:
\begin{equation}
\begin{aligned}
\IoU(H_a,G)&\geq Q_{\max}-0.02,\\
\IoU(H_a,G)&\geq Q_0+0.8(Q_{\max}-Q_0).
\end{aligned}
\end{equation}
The final training set uses one program per state. It prefers a direct teacher program within 0.005 IoU of the best retained externally supported program, otherwise choosing the highest-IoU externally supported variant. Rules supplement states without such a verified program, with their dataset share capped at 30\%.

Training annotations guide this offline search and labeling, while the reviser observes only the image, instruction, and current predicted mask.

\paragraph{Actual candidates teach Candidate Selection.}
The selection pool contains 122,876 records from executed same-target pairs, cross-target contrasts, and replay. A pair's preference follows its measured segmentation quality. Full-mask and difference views are retained so the selector can learn both target consistency and the effect of the proposed change.

Feedback Assessment uses 30,000 examples. The feedback-conditioned revision mix contains 18,000 examples, and multi-round selection adds 2,220 executed pairs with 740 replay pairs from 384 training scenes. These records can share images because they supervise different states and decisions. Predicted feedback exposes the model to its observed errors, while replay maintains previously learned revision and selection behavior.

\paragraph{Observed banks teach joint evidence use.}
For consolidation, the bank fixes region boundaries before GT assigns labels. Region labels use the exact optimum in Sec.~\ref{sec:supp_netgain}. The complete reconstruction receives IoU supervision, and reconstruction/reference pairs receive adoption targets. Initial preference proposals are sampled deterministically from an actual donor mask, its union with the reference, or their intersection. These proposals are representable in the bank's partition and are chosen independently of GT. GT evaluates the proposal afterward. This separates the source of candidate geometry from the source of its training label.

\subsection{Consolidation Loss Aligns the Decisions}
\paragraph{Mask Quality optimizes complete-mask IoU.}
The soft region inclusions are
\begin{equation}
p_i=\sigma(\ell_i).
\end{equation}
Replacing the binary inclusion bits with these probabilities gives the relaxed native-grid reconstruction
\begin{equation}
P=C+\sum_{i=1}^{K}p_iR_i.
\end{equation}
Let $S_{\mathrm{eval}}$ be the hard prediction after the official binary-mask resize, and $\mathcal R(P)$ the floating-point bilinear resize. The straight-through tensor is
\begin{equation}
P_{\mathrm{ST}}=S_{\mathrm{eval}}+
[\mathcal R(P)-\operatorname{sg}(\mathcal R(P))].
\label{eq:supp_st}
\end{equation}
Its forward value is exactly the scored hard mask. Its backward path uses the continuous relaxation~\cite{ste}. On LaSeRS the evaluation grid is $1024\times1024$, following the benchmark's uint8 bilinear-resize convention.

For each of the $N$ target states, define the intersection and union as
\begin{equation}
\begin{aligned}
I_n&=\sum_pP_{\mathrm{ST},n}(p)G_n(p),\\
U_n&=\sum_p\bigl[P_{\mathrm{ST},n}(p)+G_n(p)\bigr]-I_n.
\end{aligned}
\end{equation}
The Mask Quality loss is
\begin{equation}
\mathcal L_G=\tfrac12\left(1-\frac1N\sum_n\frac{I_n}{U_n}\right)
+\tfrac12\left(1-\frac{\sum_nI_n}{\sum_nU_n}\right).
\label{eq:supp_global_loss}
\end{equation}
Target averaging preserves small targets' influence. Pooled IoU reflects the total pixel contribution. Their combination links regional decisions to complete-mask quality~\cite{lovasz}.

Distributed training sums intersections and unions across workers before forming the pooled ratio. This preserves the objective's weighting across the full effective batch. Training averages target states, while official LaSeRS reporting follows the nine-dimension macro protocol.

\paragraph{Regional supervision emphasizes consequential evidence.}
Let $x^*$ be the native-grid optimal code and $Q^*$ its IoU. The contribution of each region at the optimum is
\begin{equation}
a_i=\tau_i-Q^*\varphi_i.
\end{equation}
The denominator scale is
\begin{equation}
D_0=\max(1,T+\varphi_0).
\end{equation}
The regional weights are
\begin{equation}
 w_i=\frac{|a_i|}{D_0+\mathbf1[x_i^*=0]\varphi_i}
 +0.01\frac{|R_i|}{\sum_j|R_j|}\mathbf1[a_i=0].
\label{eq:supp_region_weights}
\end{equation}
The first term accounts for the impact of an incorrect inclusion or omission. The second resolves equal-IoU choices toward lower background. For any inclusion code $x$, Eq.~\ref{eq:supp_netgain} gives
\begin{equation}
\begin{aligned}
Q^*-Q(x)&=\frac{\sum_i|a_i|\mathbf1[x_i\ne x_i^*]}{D(x)}\\
&\leq\sum_iw_i\mathbf1[x_i\ne x_i^*].
\end{aligned}
\label{eq:supp_regional_bound}
\end{equation}
The inequality follows because $D(x)\geq D_0$, and a falsely included region also adds its own $\varphi_i$ to the denominator. Thus, these weights bound the complete-mask cost of regional mistakes. The auxiliary objective is
\begin{equation}
\mathcal L_R=\frac{\sum_iw_i\operatorname{BCE}(\ell_i,x_i^*)}{\max(1,\sum_iw_i)}.
\label{eq:supp_region_loss}
\end{equation}
The denominator limits large auxiliary mass without amplifying tiny boundary discrepancies. Region semantics are therefore learned alongside the global quality objective, with greater emphasis on consequential decisions.

\paragraph{Relative Adoption transfers useful comparative judgments.}
For pair $j$, let $u_j$ be its measured IoU difference, $y_j=\operatorname{sign}(u_j)$, and $m_{jo}$ the reconstruction-oriented margin in order $o$. The objective is
\begin{equation}
\mathcal L_A=\frac{\sum_j|u_j|\,\frac12\sum_o
[\log(1+e^{-y_jm_{jo}})+0.5\,k_{jo}]}{\sum_j|u_j|}.
\label{eq:supp_adoption_loss}
\end{equation}
The distillation term uses temperature 2:
\begin{equation}
k_{jo}=4\,\mathrm{KL}\bigl(p_{jo}^{\mathrm{teacher}}\Vert p_{jo}^{\mathrm{student}}\bigr).
\end{equation}
It is enabled when the frozen multi-round selector's averaged preference agrees with the GT IoU ordering. This transfers the original comparison capability while keeping executed quality as the reference.

When both masks have zero target overlap, their IoUs tie. The label instead favors lower false-positive area, with utility
\begin{equation}
u_j=\frac{\mathrm{FP}_B-\mathrm{FP}_S}{|G|+\mathrm{FP}_B+\mathrm{FP}_S}.
\end{equation}
Here $\mathrm{FP}_B$ and $\mathrm{FP}_S$ count background pixels in the reference and reconstruction. This reduces background contribution to pooled error.

Zero-utility pairs contribute zero loss, and an all-zero-weight batch has zero adoption loss. The regret relation in Eq.~\ref{eq:supp_adoption_bound} applies to the non-neutral IoU pairs.

The Consolidation Loss combines the three objectives:
\begin{equation}
\mathcal L=\mathcal L_G+0.1\mathcal L_R+0.1\mathcal L_A.
\end{equation}
Regional prediction and Relative Adoption update the same consolidation adapter. The original A/B vocabulary readout is retained for preference prediction. This joins semantic region understanding, complete-mask quality, and selective adoption within one learned representation.

\subsection{Training Schedule and State Refresh}
\label{sec:supp_training_schedule}
Table~\ref{tab:supp_hyperparameters} summarizes the training hyperparameters and the two consolidation passes.
\begin{table*}[p]
\centering
\caption{\textbf{Training Hyperparameters.} Shared settings apply to all roles. LR denotes learning rate. The two consolidation passes use the LaSeRS training states described in Sec.~\ref{sec:supp_training_schedule}.}
\label{tab:supp_hyperparameters}
\small
\setlength{\tabcolsep}{6pt}
\renewcommand{\arraystretch}{1.02}
\begin{tabularx}{0.90\textwidth}{@{}L{1.5}C{0.75}C{0.75}@{}}
\toprule
\tablehead{Hyperparameter} & \multicolumn{2}{c}{\tablehead{Value}}\\
\midrule
\multicolumn{3}{l}{\textit{Shared Settings}}\\
Backbone & \multicolumn{2}{c}{Qwen3.5-2B}\\
Segmenter & \multicolumn{2}{c}{SAM3 (frozen)}\\
Optimizer & \multicolumn{2}{c}{AdamW}\\
Effective batch size & \multicolumn{2}{c}{16}\\
Weight decay & \multicolumn{2}{c}{0}\\
Warm-up ratio & \multicolumn{2}{c}{0.05}\\
Gradient clipping & \multicolumn{2}{c}{1.0}\\
LR schedule & \multicolumn{2}{c}{Cosine}\\
LoRA rank & \multicolumn{2}{c}{16}\\
LoRA alpha & \multicolumn{2}{c}{32}\\
LoRA dropout & \multicolumn{2}{c}{0}\\
\midrule
\multicolumn{3}{l}{\textit{Initial Segmentation and Selective Feedback Revision}}\\
Target Decomposition LR & \multicolumn{2}{c}{$10^{-4}$}\\
Target Decomposition epochs & \multicolumn{2}{c}{2}\\
Spatial Grounding LR & \multicolumn{2}{c}{$10^{-4}$}\\
Spatial Grounding epochs & \multicolumn{2}{c}{2}\\
Bootstrap Revision LR & \multicolumn{2}{c}{$10^{-4}$}\\
Bootstrap Revision epochs & \multicolumn{2}{c}{1}\\
Verified Revision LR & \multicolumn{2}{c}{$10^{-4}$}\\
Verified Revision epochs & \multicolumn{2}{c}{1}\\
Initial Selection LR & \multicolumn{2}{c}{$10^{-4}$}\\
Selection refinement LR (stage 1) & \multicolumn{2}{c}{$2\times10^{-5}$}\\
Selection refinement LR (stage 2) & \multicolumn{2}{c}{$10^{-5}$}\\
Selection refinement epochs & \multicolumn{2}{c}{1 per stage}\\
Feedback Assessment LR & \multicolumn{2}{c}{$5\times10^{-5}$}\\
Feedback Assessment epochs & \multicolumn{2}{c}{1}\\
Feedback-conditioned Revision LR & \multicolumn{2}{c}{$2\times10^{-5}$}\\
Multi-round Selection LR & \multicolumn{2}{c}{$10^{-5}$}\\
Random seed & \multicolumn{2}{c}{17}\\
\midrule
\multicolumn{3}{l}{\textit{Cross-Mask Evidence Consolidation}}\\
\tablehead{Hyperparameter} & \tablehead{Initial Pass} & \tablehead{Refreshed Pass}\\
\cmidrule(lr){1-3}
Adapter LR & $5\times10^{-6}$ & $2\times10^{-6}$\\
Readout LR & $3\times10^{-4}$ & $10^{-4}$\\
Final LR fraction & 0.1 & 0.1\\
Training scenes & 14,381 & 14,381\\
Training states & 43,143 & 50,343\\
Training epochs & 1 & 1\\
Training updates & 2,697 & 3,147\\
Rounds per trajectory & 3 & 3\\
Refreshed scenes & 0 & 2,400\\
Refreshed states & 0 & 7,200\\
Shuffle seed & 72 & 73\\
Region loss weight & 0.1 & 0.1\\
Adoption loss weight & 0.1 & 0.1\\
Distillation weight & 0.5 & 0.5\\
Distillation temperature & 2 & 2\\
GPUs & 4 A6000 & 4 A6000\\
\bottomrule
\end{tabularx}
\end{table*}

All adapters use LoRA rank 16, scaling parameter 32, and zero dropout. They adapt attention projections in the language backbone, including its full-attention and linear-attention layers. AdamW~\cite{adamw} uses effective batch size 16, zero weight decay, 5\% warm-up, and gradient clipping at 1. SAM3 remains frozen.

\paragraph{Initial Segmentation and selective revision.}
Target Decomposition and Spatial Grounding each use two epochs at $10^{-4}$. Bootstrap and execution-verified revision each use one epoch at $10^{-4}$, with the verified reviser trained on the backbone containing merged grounding weights.

Candidate Selection starts at $10^{-4}$, followed by one-epoch stages at $2\times10^{-5}$ and $10^{-5}$. Feedback Assessment initializes from the refined selector and uses one epoch at $5\times10^{-5}$. Feedback-conditioned revision continues at $2\times10^{-5}$, and multi-round selection adapts at $10^{-5}$ on 2,960 executed and replay pairs. These stages use cosine decay and seed 17.

\paragraph{Scene selection fixes coverage before learning.}
We use the official LaSeRS training split. A target context is eligible when its Concept Candidate has a recorded mask returned without a tool error and an executed initial-versus-revision pair is available.

The original scene partition is retained: the BLAKE2b-64 hash of the scene identifier, interpreted as an unsigned big-endian integer modulo five, assigns bucket zero to development and the other four buckets to training. Within each training scene, we select the eligible context with the smallest BLAKE2b-128 hash of its state identifier. The context preserves the complete instruction and selected target.

This yields 14,381 distinct training scenes, with one target context per scene. Selection depends on recorded execution availability and identifiers, not segmentation quality, giving each scene equal representation.

\paragraph{Consolidation learns from complete trajectories.}
The consolidation adapter is initialized with the multi-round selector's parameters. Its new regional readout and frozen anchor have identical parameters. Each selected context supplies three round-specific states, for 43,143 states. The initial learning pass uses learning rates of $5\times10^{-6}$ for the adapter and $3\times10^{-4}$ for the readout, with 2,697 updates.

For refresh, the selected training contexts are ranked by the first 64 bits of SHA-256 applied to the fixed prefix \texttt{v72\_refresh} followed by the state identifier. The first 2,400 form the refresh set, fixed before training and retaining the same target contexts.

After the initial learning pass, the current model re-executes their three-round trajectories. Their 7,200 new states augment the original pool to 50,343 states. A second pass uses $2\times10^{-6}$ and $10^{-4}$, respectively, for 3,147 updates. Both passes use four A6000 GPUs, the Consolidation Loss, and cosine decay to one-tenth of the peak learning rate. Shuffle seeds are 72 and 73. The second pass retains the first pass's anchor.

State refresh adapts the model to the trajectories induced by its own decisions, following the principle of learning on model-induced states~\cite{dagger,gkd}. Upstream roles stay frozen during consolidation training, while their inference states respond to the adopted mask. Thus, new compositions participate in subsequent feedback rather than being appended only at the end.

\section{Evaluation Protocols and Metric Definitions}
\label{sec:supp_eval}
\subsection{Benchmarks and Baselines}
\label{sec:supp_baselines}
Each benchmark uses its corresponding training split and a separately trained GeoReVis model. Test scenes are excluded from supervision construction and state refresh. LaSeRS uses its official macro metrics across nine dimensions. Query-level insight statistics, target-level capacity statistics, and ReVis-Eval item averages use their explicitly stated units rather than the macro average across these dimensions.

Retrained baselines use the corresponding benchmark's training split and retain the training objectives, optimizer settings, and schedules of their public implementations. The $^\ddagger$ variants retain their native architectures. The $^\ast$ variants use Qwen3.5-2B with LoRA and frozen SAM3, adapting the model and spatial-prompt interfaces while retaining each method's learning objective. All feedback-revision models are evaluated for three revision rounds. SAM~3 Agent is evaluated zero-shot.

For LISA, PixelLM, GLaMM-ft, M$^2$A, GeoPixel, and SegEarth-R2, the LaSeRS results are taken from SegEarth-R2~\cite{segearthr2}.

\paragraph{LISA.}
We retain LLaVA-7B, the segmentation-token interface, and the SAM ViT-H decoder~\cite{lisa,sam}. Retraining follows the released supervised fine-tuning recipe, jointly optimizing language outputs and predicted masks on the target benchmark.

\paragraph{PixelLM.}
We use LLaVA-7B with PixelLM's segmentation codebook and lightweight pixel decoder~\cite{pixellm}. The released language and mask supervision is retained, with target-benchmark annotations converted into its text-and-mask training format.

\paragraph{GeoPixel.}
We retain the IXC-2.5-7B backbone, SAM2-L~\cite{sam2}, and the high-resolution image-processing pipeline~\cite{geopixel}. Training follows the released pixel-grounding objective and fine-tuning configuration using the corresponding benchmark's instructions and masks.

\paragraph{SegEarth-R1.}
We retain its language-conditioned mask-generation architecture and released supervised training recipe~\cite{segearthr1}. The RISBench reproduction uses that benchmark's training split with its image--instruction--mask annotations, while the other reported entries are taken from published results.

\paragraph{RMSIN.}
We report published referring segmentation results for the original Swin/BERT-based network with rotated multi-scale interaction modules~\cite{rmsin}. RMSIN learns through supervised mask prediction using separate visual and language encoders rather than a generative VLM.

\paragraph{RS2-SAM 2.}
We retain the joint image--text encoder, hierarchical fusion modules, and dense-prompt generator for SAM2~\cite{rs2sam2}. The RISBench reproduction uses the released supervised segmentation objective. The model predicts prompts from aligned features rather than autoregressive VLM outputs.

\paragraph{RSRefSeg 2.}
We report published referring segmentation results for the contrastive vision--language encoder, SAM-based decoder, and learned prompting modules of RSRefSeg~2~\cite{rsrefseg2}. Its parameter-efficient adaptation uses mask supervision and feature-based language grounding.

\paragraph{CROSS.}
We retain SigLIP~2 and SAM~2 with the cascaded distillation and dual-constraint grounding modules~\cite{cross}. RISBench training follows the published combination of segmentation, distillation, and spatial contrastive losses, keeping the original parameter-efficient adaptation design.

\paragraph{RemoteReasoner.}
We retain the Qwen2.5-VL-7B reasoning model and its segmentation-tool interface~\cite{remotereasoner}. Retraining follows the released GRPO recipe on the corresponding target-benchmark annotations, preserving its spatial-prompt generation and task-transformation strategy.

\paragraph{Seg-Zero.}
We use Qwen2.5-VL-7B with frozen SAM2-L and retrain on LaSeRS~\cite{segzero}. The released GRPO objective retains format and spatial-accuracy rewards for prompt generation. Inference produces a segmentation from the predicted prompts in a single pass.

\paragraph{VisionReasoner.}
The native reproduction uses Qwen2.5-VL-7B and SAM2-L with the released multi-object GRPO training recipe~\cite{visionreasoner}. VisionReasoner$^\ast$ replaces these components with Qwen3.5-2B and frozen SAM3. Both retain single-pass spatial prompting and the original reward formulation.

\paragraph{Think2Seg-RS.}
We use the Qwen2.5-VL-7B variant with frozen SAM2-S~\cite{think2segrs}. Training follows the released GRPO objective, combining output-format rewards with executed-mask IoU rewards on the corresponding benchmark. The VLM is optimized while the segmentation module remains fixed.

\paragraph{SegEarth-R2.}
We report the published SegEarth-R2 results on all five benchmarks~\cite{segearthr2}. The original study trains separately on each benchmark's official training split. Its referring segmentation comparison reports gIoU, so the corresponding unreported cIoU entries are left blank in Table~\ref{tab:supp_referring}.

\paragraph{UniGeoSeg.}
We retain Phi-1.5, the Swin-B visual encoder, and Mask2Former, together with task-aware instruction encoding and latent knowledge memory~\cite{unigeoseg}. Retraining on RefSegRS and RISBench follows the released segmentation-supervision and optimization settings.

\paragraph{GeoReason.}
We report the published EarthReason and RRSIS-D results of GeoReason~\cite{georeason}. It uses Qwen3-VL-4B and SAM2 initialized from Sa2VA~\cite{sa2va}, with auxiliary remote sensing VQA and caption data from GeoChatInstruct~\cite{geochat} and VRSBench~\cite{vrsbench}. Its training combines language and mask supervision with latent patch-to-token alignment. For LaSeRS, RefSegRS, and RISBench, we train GeoReason separately on each benchmark's corresponding training split. These reproduced results are marked with $^\ddagger$.

\paragraph{SegAgent.}
The native version uses Qwen-VL-7B and SAM ViT-H~\cite{segagent}. Training follows its trajectory-imitation recipe for corrective point prompting from mask observations. The SegAgent$^\ast$ variant used in ReVis-Eval adopts Qwen3.5-2B and frozen SAM3 while retaining the same trajectory-learning objective.

\paragraph{SAM-Veteran.}
We retain its multi-task GRPO recipe for box grounding, mask-aware point refinement, and termination~\cite{samveteran}. The native version uses Qwen2.5-VL-7B and SAM2-L. SAM-Veteran$^\ast$ uses Qwen3.5-2B and frozen SAM3, with the original refinement actions and reward design.

\paragraph{SocioReasoner.}
We retain the two-step training objective: localization rewards supervise initial boxes, and executed-mask rewards supervise feedback-conditioned refinement~\cite{socioreasoner}. The native version uses Qwen2.5-VL-3B and SAM2-L, while SocioReasoner$^\ast$ uses Qwen3.5-2B and frozen SAM3. Both receive the optical image and instruction without a digital map. The three segmenter calls correspond to three revision rounds, excluding initial segmentation.

\subsection{ReVis-Eval Protocol Construction}
\paragraph{Revision Degradation Evaluation.}
The segmentation pipelines of GeoReVis, VisionReasoner$^\ddagger$, and SocioReasoner$^\ddagger$ each generate one initial mask per test query without revision. Removing initial masks with IoU above 95\% leaves 3,201 of 3,474 items. All tested methods receive the same image, instruction, and initial mask and perform three revision rounds. Feedback Gain uses a paired run with the same budget and views, replacing the mask overlay with the plain image. This comparison measures the contribution of mask feedback within each method.

\paragraph{Evidence Fragmentation Evaluation.}
GeoReVis revision and selection, SAM-Veteran$^\ast$, and SocioReasoner$^\ast$ each perform three revision rounds to generate trajectories. The resulting 3,474 banks retain the final committed reference mask $B$ and all executed candidate masks. Banks contain \mbox{4-10} masks, averaging 6.4. Methods may output pixels or combine existing masks through union, intersection, difference, and cropping, without new segmenter calls. The 812-bank high-potential subset $P$ has a regional oracle exceeding the best complete mask by more than 5\% IoU. Generators, seeds, tool outputs, and geometry are fixed for each item.

\subsection{Process Evaluation Metrics}
\label{sec:supp_process_metrics}
For evaluation, masks are treated as sets of foreground pixels. For prediction $M$ and ground truth $G$, IoU is
\begin{equation}
\IoU(M,G)=\frac{|M\cap G|}{|M\cup G|}.
\end{equation}
Here $|\cdot|$ denotes set size: pixel count for masks and item count for index sets. The operations $\cap$, $\cup$, and $\setminus$ denote intersection, union, and set difference.

Each evaluation item combines an image and instruction with an initial mask or a candidate bank. $G_i$ is the ground-truth target mask for item $i$. All metrics average items equally. IoU is computed on the $[0,1]$ scale, and tables multiply the scores below by 100. IoU differences denote absolute changes. Scores, differences, and rates use \% notation after scaling by 100.

\paragraph{Revision Degradation Evaluation.}
There are $N_R=3{,}201$ initial-mask items. Let $M_{0,i}$ be the initial mask and $M_{T,i}$ the final mask after $T=3$ rounds. The paired output $M^{\mathrm{plain}}_{T,i}$ is obtained by replacing mask overlays with plain images under the same evaluation settings and budget. Their IoU scores are
\begin{equation}
\begin{aligned}
q_{0,i}&=\IoU(M_{0,i},G_i),\\
q_{T,i}&=\IoU(M_{T,i},G_i),\\
q^{\mathrm{plain}}_{T,i}&=\IoU(M^{\mathrm{plain}}_{T,i},G_i).
\end{aligned}
\end{equation}
\begin{enumerate}[label=\textbf{(\arabic*)},leftmargin=*,itemsep=3pt,topsep=3pt,parsep=0pt]
\item \textbf{IoU Gain ($\Delta$IoU).} For each item, subtract initial IoU from final IoU, then average:
\begin{equation}
 \Delta\mathrm{IoU}=\frac{1}{N_R}\sum_{i=1}^{N_R}
 (q_{T,i}-q_{0,i}).
\end{equation}
Positive values indicate net segmentation improvement across the evaluated items.

\item \textbf{Feedback Gain (FG).} Compare the final masks from the feedback run and its paired plain-image run:
\begin{equation}
 \mathrm{FG}=\frac{1}{N_R}\sum_{i=1}^{N_R}
 (q_{T,i}-q^{\mathrm{plain}}_{T,i}).
\end{equation}
Higher FG indicates greater improvement from observing the mask under the matched budget.

\item \textbf{Degradation Rate (DR).} Count items whose final IoU falls more than 0.01 below their initial IoU:
\begin{equation}
 \mathrm{DR}=\frac{1}{N_R}\sum_{i=1}^{N_R}
 \mathbf{1}[q_{T,i}-q_{0,i}<-0.01].
\end{equation}
The indicator $\mathbf{1}[\cdot]$ equals one when its condition holds and zero otherwise. The threshold is 0.01 IoU. Lower DR indicates fewer degraded outputs.

\item \textbf{Correct-Pixel Preservation (CP).} Define the initially correct target pixels and the items containing them as
\begin{align}
C_i^0&=M_{0,i}\cap G_i,\\
\mathcal I^R_{\mathrm{CP}}&=\{i:|C_i^0|>0\}.
\end{align}
Their preservation score is
\begin{equation}
 \mathrm{CP}=\frac{1}{|\mathcal I^R_{\mathrm{CP}}|}
 \sum_{i\in\mathcal I^R_{\mathrm{CP}}}\frac{|M_{T,i}\cap C_i^0|}{|C_i^0|}.
\end{equation}
The numerator counts initially correct target pixels retained in the final mask. Higher CP indicates better preservation of correct regions.
\end{enumerate}

\paragraph{Evidence Fragmentation Evaluation.}
There are $N_E=3{,}474$ candidate-bank items. For item $i$, $\mathcal H_i$ contains the candidate masks, $B_i$ is the trajectory's final committed reference mask, and $F_i$ is the output of the evaluated method. The optimal candidate is determined by GT IoU for scoring.
\begin{enumerate}[label=\textbf{(\arabic*)},leftmargin=*,itemsep=3pt,topsep=3pt,parsep=0pt]
\item \textbf{Best-Whole Gap (BWG).} Let $q_i^\star$ be the IoU of the optimal candidate and $d_i$ the output's gain over it:
\begin{align}
q_i^\star&=\max_{H\in\mathcal H_i}\IoU(H,G_i),\\
d_i&=\IoU(F_i,G_i)-q_i^\star.
\end{align}
BWG averages this gain across all banks:
\begin{equation}
\mathrm{BWG}=\frac{1}{N_E}\sum_{i=1}^{N_E}d_i.
\end{equation}
Positive BWG indicates average improvement over the GT-best single candidate. Selecting one existing candidate cannot yield a positive per-item gap.

\item \textbf{BWG on High-Potential Banks (BWG$_P$).} Let $q_i^{\mathrm{reg}}$ be the IoU of the GT-selected composition under the bank's fixed regional partition. The high-potential subset is
\begin{equation}
P=\{i:q_i^{\mathrm{reg}}-q_i^\star>0.05\}.
\end{equation}
It contains 812 banks under the 0.05-IoU threshold. Its score is
\begin{equation}
\mathrm{BWG}_P=\frac{1}{|P|}\sum_{i\in P}d_i.
\end{equation}
Higher BWG$_P$ indicates better use of the available consolidation potential. The remaining banks form
\begin{equation}
\bar P=\{1,\ldots,N_E\}\setminus P.
\end{equation}
BWG$_{\bar P}$ applies the same average to this subset.

\item \textbf{Historical Recall (HR).} Let $U_i$ be the union of candidates and $E_i$ the target pixels present in history but missing from the reference:
\begin{equation}
\begin{aligned}
U_i&=\bigcup_{H\in\mathcal H_i}H,\\
E_i&=(G_i\cap U_i)\setminus B_i,\\
\mathcal I_{\mathrm{HR}}&=\{i:|E_i|>0\}.
\end{aligned}
\end{equation}
We average recall over items containing such evidence:
\begin{equation}
 \mathrm{HR}=\frac{1}{|\mathcal I_{\mathrm{HR}}|}
 \sum_{i\in\mathcal I_{\mathrm{HR}}}\frac{|F_i\cap E_i|}{|E_i|}.
\end{equation}
The numerator counts recovered target pixels and the denominator counts those available for recovery. Higher HR indicates more complete recovery of historical evidence.

\item \textbf{Added Precision (AP).} Define the newly included pixels and the items with nonempty additions as
\begin{align}
A_i&=F_i\setminus B_i,\\
\mathcal I_{\mathrm{AP}}&=\{i:|A_i|>0\}.
\end{align}
Their precision is
\begin{equation}
 \mathrm{AP}=\frac{1}{|\mathcal I_{\mathrm{AP}}|}
 \sum_{i\in\mathcal I_{\mathrm{AP}}}\frac{|A_i\cap G_i|}{|A_i|}.
\end{equation}
The numerator counts correct additions and the denominator counts all additions. Higher AP indicates more precise evidence recovery.
\end{enumerate}
For CP, HR, and AP, the index sets exclude items with zero denominators. Such items are omitted from that metric's average, rather than assigned a score of zero or one. These are averages of per-item ratios, not ratios of pooled pixel counts. HR and AP use the committed reference $B_i$, while BWG uses the GT-optimal candidate.

\paragraph{Additional diagnostics separate error reduction and evidence retention.}
The complete revision results also report Net Error Reduction (NER). Define the initial and final error pixels as
\begin{align}
E_i^0&=M_{0,i}\triangle G_i,\\
E_i^T&=M_{T,i}\triangle G_i,
\end{align}
where $\triangle$ is symmetric difference. Their relative reduction is
\begin{equation}
 \mathrm{NER}=\frac{1}{N_R}\sum_{i=1}^{N_R}
 \left(1-\frac{|E_i^T|}{|E_i^0|}\right).
\end{equation}
Positive NER indicates a reduction in pixel errors.

For evidence consolidation, define the correct reference pixels and their valid items as
\begin{align}
C_i^B&=B_i\cap G_i,\\
\mathcal I^E_{\mathrm{CP}}&=\{i:|C_i^B|>0\}.
\end{align}
The supplementary CP score is
\begin{equation}
 \mathrm{CP}_E=\frac{1}{|\mathcal I^E_{\mathrm{CP}}|}
 \sum_{i\in\mathcal I^E_{\mathrm{CP}}}\frac{|F_i\cap C_i^B|}{|C_i^B|}.
\end{equation}
This score measures preservation of the reference mask's correct target pixels. Taking the union of all candidates yields HR and $\mathrm{CP}_E$ of one on valid items, but can introduce background, so AP and final IoU are evaluated alongside recovery.

\section{Extended Experimental Evidence}
\label{sec:supp_results}
\subsection{Complete Segmentation Results}
\label{sec:supp_full_results}
\begin{figure}[!t]
\centering\includegraphics[width=\linewidth]{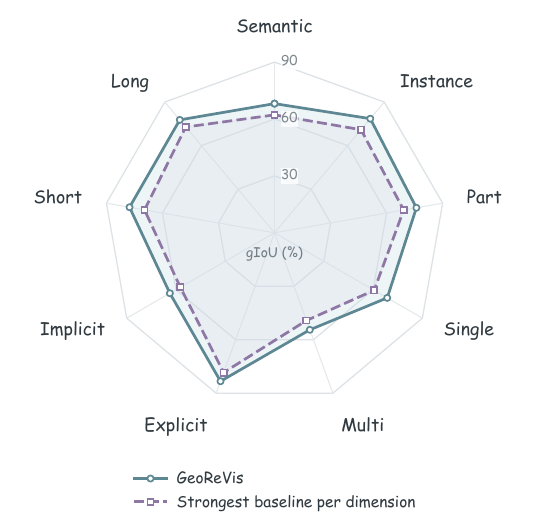}
\caption{\textbf{Performance across nine LaSeRS dimensions.} All axes report gIoU (\%) on the same 0--90 scale. The baseline profile takes the highest non-GeoReVis score in each dimension of main Table~1. GeoReVis exceeds this profile on all nine axes.}
\label{fig:supp_dimensions}
\end{figure}

Table~\ref{tab:supp_lasers} provides gIoU and cIoU for all nine LaSeRS dimensions. Figure~\ref{fig:supp_dimensions} compares GeoReVis with the strongest baseline across the nine gIoU dimensions. GeoReVis improves every dimension, with gains spanning 4.89\% to 8.02\%. Its advantage covers small parts, multiple targets, and implicit instructions, matching the settings in which selective revision and evidence consolidation are most useful.
\begin{table*}[!t]\centering
\caption{Full LaSeRS results (gIoU/cIoU, \%) for every dimension. Backbones and segmenters follow Table~1 of the main paper.}\label{tab:supp_lasers}
\scriptsize\setlength{\tabcolsep}{2.2pt}
\resizebox{\linewidth}{!}{\begin{tabular}{lcccccccccc}\toprule
\tablehead{Method} & \tablehead{Semantic} & \tablehead{Instance} & \tablehead{Part} & \tablehead{Single} & \tablehead{Multi} & \tablehead{Explicit} & \tablehead{Implicit} & \tablehead{Short} & \tablehead{Long} & \tablehead{Mean}\\\midrule
\multicolumn{11}{c}{\textbf{\textit{Single-pass Segmentation Models}}}\\\midrule
LISA~\cite{lisa} & 26.40/23.20 & 20.50/25.00 & 16.10/11.60 & 37.30/32.20 & 18.20/22.40 & 27.10/24.30 & 21.50/25.60 & 34.10/27.80 & 38.40/33.90 & 26.60/25.10\\
PixelLM~\cite{pixellm} & 32.00/32.80 & 26.60/30.00 & 13.20/16.50 & 44.30/40.40 & 20.20/23.50 & 25.00/23.10 & 23.90/21.90 & 41.60/38.90 & 37.10/34.50 & 29.30/29.10\\
M$^2$A~\cite{mmr} & 30.10/33.00 & 23.00/24.80 & 18.60/17.20 & 45.40/37.60 & 20.90/24.80 & 35.80/30.40 & 23.30/26.70 & 35.80/32.80 & 41.50/36.70 & 30.50/29.30\\
GLaMM-ft~\cite{glamm} & 44.80/47.90 & 41.20/48.30 & 32.60/42.70 & 47.30/50.30 & 32.20/41.00 & 59.10/60.30 & 42.60/44.80 & 50.40/54.80 & 42.60/44.80 & 43.60/48.30\\
GeoPixel~\cite{geopixel} & 51.40/57.20 & 44.10/49.30 & 43.90/52.40 & 55.00/45.80 & 49.20/49.70 & 66.50/61.30 & 41.10/58.30 & 51.10/59.30 & 51.40/63.20 & 50.40/55.20\\
Seg-Zero$^\ddagger$~\cite{segzero} & 47.37/52.31 & 60.60/59.54 & 57.84/64.60 & 48.64/47.26 & 31.16/45.86 & 66.76/71.28 & 43.16/61.38 & 60.23/62.48 & 59.18/61.79 & 52.77/58.50\\
VisionReasoner$^\ddagger$~\cite{visionreasoner} & 48.98/54.83 & 62.05/60.47 & 59.52/65.61 & 49.15/47.49 & 33.31/48.32 & 67.22/72.16 & 44.31/62.31 & 60.62/63.54 & 61.12/63.45 & 54.03/59.80\\
Think2Seg-RS$^\ddagger$~\cite{think2segrs} & 52.08/59.13 & 65.37/65.16 & 60.85/71.07 & 52.11/54.54 & 37.72/54.92 & 69.78/78.34 & 46.70/68.49 & 64.05/68.41 & 63.88/69.61 & 56.95/65.52\\
SegEarth-R2~\cite{segearthr2} & 60.20/71.80 & 65.40/70.30 & 64.80/68.30 & 55.10/69.20 & 38.30/56.20 & 78.40/80.40 & 42.80/59.70 & 60.20/69.90 & 50.10/65.70 & 57.20/67.90\\
VisionReasoner$^\ast$ & 60.71/59.87 & 70.62/64.27 & 67.42/68.85 & 59.85/54.70 & 45.48/54.82 & 75.78/76.20 & 55.48/66.09 & 69.58/66.87 & 70.55/67.73 & 63.94/64.38\\
GeoReason$^\ddagger$~\cite{georeason} & 62.18/64.51 & 70.96/70.09 & 69.14/72.68 & 60.62/59.83 & 46.86/58.47 & 75.48/79.36 & 57.36/69.72 & 69.24/71.18 & 72.81/72.61 & 64.96/68.72\\
\midrule
\multicolumn{11}{c}{\textbf{\textit{Feedback-revision Segmentation Models}}}\\\midrule
SAM~3 Agent~\cite{sam3} & 37.50/39.85 & 52.49/47.27 & 48.33/52.63 & 38.34/33.79 & 20.08/33.92 & 58.44/60.79 & 32.32/49.24 & 51.33/51.13 & 50.51/51.14 & 43.26/46.64\\
SegAgent$^\ddagger$~\cite{segagent} & 50.37/56.13 & 63.32/63.01 & 60.64/68.18 & 51.93/51.86 & 33.64/50.22 & 68.18/74.34 & 45.88/65.14 & 61.81/64.66 & 62.12/66.18 & 55.32/62.19\\
SAM-Veteran$^\ddagger$~\cite{samveteran} & 53.34/58.22 & 66.12/65.65 & 61.62/70.15 & 53.32/52.56 & 38.19/52.86 & 70.93/75.29 & 48.15/66.45 & 63.77/66.72 & 63.23/67.48 & 57.63/63.93\\
SocioReasoner$^\ddagger$~\cite{socioreasoner} & 55.73/61.80 & 67.04/67.19 & 63.99/72.75 & 55.13/57.06 & 39.85/57.13 & 70.79/79.24 & 49.77/69.38 & 65.69/71.13 & 65.99/70.37 & 59.33/67.34\\
SAM-Veteran$^\ast$ & 59.39/63.40 & 69.20/69.45 & 66.78/73.04 & 58.48/58.95 & 44.26/59.12 & 74.52/79.15 & 54.98/70.98 & 67.75/70.26 & 69.03/71.44 & 62.71/68.42\\
SocioReasoner$^\ast$ & 58.86/63.48 & 69.78/70.81 & 67.70/73.61 & 59.86/58.70 & 45.71/59.96 & 74.29/80.22 & 54.16/70.44 & 69.61/70.91 & 68.20/72.60 & 63.13/68.97\\
GeoReVis w/o Consolidation & 64.80/68.95 & 75.79/74.40 & 72.52/77.85 & 65.64/63.64 & 52.13/63.56 & 81.23/84.94 & 60.53/74.88 & 74.74/76.25 & 75.33/76.37 & 69.19/73.43\\
\rowcolor{paperteal!9}\textbf{GeoReVis (ours)} & 68.11/72.41 & 78.56/77.72 & 75.95/81.58 & 68.60/67.54 & 54.42/67.43 & 83.29/87.59 & 63.73/79.16 & 77.63/79.70 & 77.73/79.91 & 72.00/77.01\\
\bottomrule
\end{tabular}}\end{table*}

Tables~\ref{tab:supp_earth} and~\ref{tab:supp_referring} give the complete reasoning and referring segmentation comparisons. GeoReVis leads all four EarthReason metrics and the gIoU scores of all three referring benchmarks. These results extend its effectiveness from implicit requests to explicitly specified target regions. The largest overall advantage occurs on LaSeRS, whose part-level and multi-target instructions place greater demands on preserving and reusing local evidence.
\begin{table}[!htbp]\centering
\caption{Full reasoning segmentation results on EarthReason (\%).}\label{tab:supp_earth}
\footnotesize\setlength{\tabcolsep}{3pt}
\begin{tabularx}{\linewidth}{Xcccc}\toprule
\multicolumn{1}{l}{\multirow{2}{*}[-0.4ex]{\tablehead{Method}}} & \multicolumn{2}{c}{\tablehead{Val}} & \multicolumn{2}{c}{\tablehead{Test}}\\
\cmidrule(lr){2-3}\cmidrule(lr){4-5}
& \tablehead{gIoU} & \tablehead{cIoU} & \tablehead{gIoU} & \tablehead{cIoU}\\\midrule
GeoPixel~\cite{geopixel} & 52.13 & 54.23 & 52.53 & 53.90\\
PixelLM~\cite{pixellm} & 57.94 & 57.79 & 60.01 & 59.22\\
LISA~\cite{lisa} & 61.04 & 57.39 & 60.88 & 59.10\\
VisionReasoner~\cite{visionreasoner} & 66.58 & 67.13 & 67.19 & 65.41\\
SegEarth-R1~\cite{segearthr1} & 68.60 & 64.13 & 70.75 & 68.25\\
RemoteReasoner~\cite{remotereasoner} & 69.02 & 67.80 & 70.96 & 69.13\\
SegEarth-R2~\cite{segearthr2} & 72.30 & 68.10 & 73.50 & 69.50\\
UniGeoSeg~\cite{unigeoseg} & 72.54 & 73.30 & 73.08 & 74.61\\
Think2Seg-RS~\cite{think2segrs} & 72.16 & 72.87 & 73.36 & 75.60\\
VisionReasoner$^\ast$~\cite{visionreasoner} & 74.36 & 75.71 & 74.57 & 75.29\\
SocioReasoner$^\ast$~\cite{socioreasoner} & 74.07 & 74.91 & 75.32 & 76.18\\
GeoReason~\cite{georeason} & \underline{77.61} & \underline{77.63} & \underline{78.61} & \underline{80.60}\\
\rowcolor{paperteal!9}\textbf{GeoReVis (ours)} & \textbf{77.63} & \textbf{79.35} & \textbf{79.28} & \textbf{81.17}\\
\bottomrule\end{tabularx}\end{table}

\begin{table}[!htbp]\centering
\caption{Full referring segmentation results (test gIoU/cIoU, \%).}\label{tab:supp_referring}
\footnotesize\setlength{\tabcolsep}{1.5pt}
\begin{tabularx}{\linewidth}{Xccccc}\toprule
\multicolumn{1}{l}{\multirow{2}{*}[-0.4ex]{\tablehead{Method}}} & \multicolumn{2}{c}{\tablehead{RRSIS-D}} & \multicolumn{2}{c}{\tablehead{RefSegRS}} & \tablehead{RISBench}\\
\cmidrule(lr){2-3}\cmidrule(lr){4-5}\cmidrule(lr){6-6}
& \tablehead{gIoU} & \tablehead{cIoU} & \tablehead{gIoU} & \tablehead{cIoU} & \tablehead{gIoU}\\\midrule
LISA$^\ddagger$~\cite{lisa} & 26.78 & 27.94 & 24.62 & 35.17 & 28.36\\
PixelLM$^\ddagger$~\cite{pixellm} & 31.65 & 33.60 & 28.95 & 39.48 & 32.47\\
RMSIN~\cite{rmsin} & 64.20 & 77.79 & 62.58 & 75.72 & 63.07\\
VisionReasoner$^\ddagger$~\cite{visionreasoner} & 63.47 & 73.85 & 69.71 & 76.93 & 66.41\\
RemoteReasoner$^\ddagger$~\cite{remotereasoner} & 64.12 & 74.36 & 71.28 & 77.64 & 66.95\\
SocioReasoner$^\ast$~\cite{socioreasoner} & 64.61 & 74.92 & 71.94 & 78.02 & 67.62\\
GeoPixel$^\ddagger$~\cite{geopixel} & 67.30 & \textbf{84.90} & 70.36 & 78.45 & 66.83\\
Think2Seg-RS$^\ddagger$~\cite{think2segrs} & 65.08 & 75.21 & 72.85 & 78.41 & 68.27\\
SegEarth-R1$^\ddagger$~\cite{segearthr1} & 66.40 & 78.01 & 72.45 & 79.00 & 68.12\\
RS2-SAM 2$^\ddagger$~\cite{rs2sam2} & 66.72 & 78.99 & 73.90 & 80.87 & 68.55\\
SegEarth-R2~\cite{segearthr2} & 67.90 & -- & 74.80 & -- & 70.50\\
GeoReason$^\ddagger$~\cite{georeason} & 67.32 & 79.39 & 76.38 & 81.24 & 71.36\\
UniGeoSeg$^\ddagger$~\cite{unigeoseg} & \underline{69.25} & 79.78 & 76.06 & 81.02 & 70.94\\
RSRefSeg 2~\cite{rsrefseg2} & 69.17 & 79.45 & 77.39 & 81.24 & \underline{72.57}\\
CROSS$^\ddagger$~\cite{cross} & 68.92 & 79.89 & \underline{79.51} & \underline{83.25} & 70.86\\
\rowcolor{paperteal!9}\textbf{GeoReVis (ours)} & \textbf{70.35} & \underline{80.47} & \textbf{80.72} & \textbf{84.11} & \textbf{74.27}\\
\bottomrule\end{tabularx}\end{table}

Table~\ref{tab:supp_er_ablation} further connects EarthReason performance to the method components. Selective revision and multi-round feedback improve the initial prediction, while consolidation adds 3.91\% gIoU and 3.93\% cIoU. The corresponding LaSeRS gains are 2.81\% and 3.58\%. Evidence consolidation therefore benefits both benchmark-specific training settings.
\begin{table}[!htbp]\centering
\caption{Component analysis on EarthReason test (gIoU/cIoU). The one-round row includes the Concept Candidate.}\label{tab:supp_er_ablation}
\footnotesize\setlength{\tabcolsep}{3pt}
\begin{tabular}{lcc}\toprule
\tablehead{Configuration} & \tablehead{gIoU} & \tablehead{cIoU}\\\midrule
Initial Segmentation & 71.46 & 70.38\\
Selective Revision (1 Round) & 74.21 & 76.02\\
+ Multi-Round Feedback & 75.37 & 77.24\\
\rowcolor{paperteal!9}+ Evidence Consolidation & 79.28 & 81.17\\
\bottomrule\end{tabular}\end{table}

\subsection{Component and Supervision Ablations}
\label{sec:supp_component}
Table~\ref{tab:supp_full_ablation} extends main Table~6 with 254 fragmentation-heavy queries. Cross-Mask Evidence Consolidation raises their gIoU from 56.23\% to 60.69\%. This 4.46\% gain exceeds its 2.81\% gain across LaSeRS. The improvement is concentrated where recovering complementary evidence is especially valuable.
\begin{table*}[!t]\centering
\caption{\textbf{Component Ablation on LaSeRS.} U.R.: Unverified Revision. V.R.: Verified Revision. C.S.: Candidate Selection. C.C.: Concept Candidate. M.F.: Multi-Round Feedback. E.C.: Evidence Consolidation. U.R. and V.R. are alternative revisions trained on unverified and execution-verified actions. The first row is Initial Segmentation, and the last is full GeoReVis. Frag.: gIoU on the fixed 254-query fragmentation-heavy subset.}\label{tab:supp_full_ablation}
\small\setlength{\tabcolsep}{4pt}\renewcommand{\arraystretch}{1.10}
\begin{tabularx}{\linewidth}{*{6}{C{0.9}}*{5}{C{1.12}}}\toprule
\multicolumn{6}{c}{\tablehead{Components}} & \multicolumn{5}{c}{\tablehead{LaSeRS}}\\
\cmidrule(lr){1-6}\cmidrule(lr){7-11}
\tablehead{U.R.} & \tablehead{V.R.} & \tablehead{C.S.} & \tablehead{C.C.} & \tablehead{M.F.} & \tablehead{E.C.} & \tablehead{gIoU} & \tablehead{cIoU} & \tablehead{DR$\downarrow$} & \tablehead{OR$\downarrow$} & \tablehead{Frag.}\\\midrule
$\times$ & $\times$ & $\times$ & $\times$ & $\times$ & $\times$ & 63.28 & 62.85 & -- & -- & 51.36\\
$\checkmark$ & $\times$ & $\times$ & $\times$ & $\times$ & $\times$ & 62.71 & 62.93 & 32.47 & 37.05 & 50.28\\
$\times$ & $\checkmark$ & $\times$ & $\times$ & $\times$ & $\times$ & 64.64 & 66.65 & 27.81 & 36.44 & 51.02\\
$\times$ & $\checkmark$ & $\checkmark$ & $\times$ & $\times$ & $\times$ & 66.58 & 70.62 & 10.36 & 33.16 & 54.47\\
$\times$ & $\checkmark$ & $\checkmark$ & $\checkmark$ & $\times$ & $\times$ & 67.90 & 72.81 & 8.72 & 35.92 & 54.72\\
$\times$ & $\checkmark$ & $\checkmark$ & $\checkmark$ & $\checkmark$ & $\times$ & 69.19 & 73.43 & 7.34 & 38.69 & 56.23\\
\rowcolor{paperteal!9}$\times$ & $\checkmark$ & $\checkmark$ & $\checkmark$ & $\checkmark$ & $\checkmark$ & \textbf{72.00} & \textbf{77.01} & \textbf{6.13} & \textbf{27.46} & \textbf{60.69}\\
\bottomrule\end{tabularx}
\end{table*}

\paragraph{Execution-driven labels support the complete learning pipeline.}
Table~\ref{tab:supp_training} compares supervision sources under single-round revision and selection. Verified actions and executed preferences reach 67.90\% gIoU, improving on unverified teacher actions by 1.20\% and synthetic perturbation pairs by 0.86\%. The same principle extends to consolidation: candidate geometry comes from actual outputs, while GT assigns region, quality, and adoption targets.
\begin{table}[!htbp]\centering
\caption{\textbf{Execution-Driven Supervision and Joint Learning.} LaSeRS gIoU/cIoU (\%). Top: consolidation objectives. Middle: supervision sources for single-round revision and selection, including the concept-prompt candidate. Bottom: loss weights and retained history.}\label{tab:supp_training}
\footnotesize\setlength{\tabcolsep}{3pt}
\begin{tabular}{lcc}\toprule
\tablehead{Setting} & \tablehead{gIoU} & \tablehead{cIoU}\\\midrule
\rowcolor{paperteal!9}Full objective & 72.00 & 77.01\\
w/o whole-mask loss $\mathcal L_G$ & 71.71 & 76.58\\
$\mathcal L_G$: target-average term only & 72.04 & 76.67\\
$\mathcal L_G$: pooled term only & 71.86 & 77.12\\
w/o region loss $\mathcal L_R$ & 71.93 & 76.95\\
w/o adoption loss $\mathcal L_A$ & 71.82 & 76.87\\
w/o student-state refresh & 71.89 & 76.83\\
\midrule
Verified actions + executed pairs & 67.90 & 72.81\\
Unverified teacher actions & 66.70 & 71.65\\
Rule-generated actions only & 67.15 & 72.36\\
Synthetic perturbation pairs & 67.04 & 71.80\\
\midrule
$\lambda_R=\lambda_A=0.05$ & 71.97 & 76.94\\
\rowcolor{paperteal!9}$\lambda_R=\lambda_A=0.1$ (default) & 72.00 & 77.01\\
$\lambda_R=\lambda_A=0.2$ & 71.90 & 76.99\\
History: current round & 71.44 & 76.32\\
History: last two rounds & 71.96 & 76.97\\
\rowcolor{paperteal!9}History: all rounds (default) & 72.00 & 77.01\\
\bottomrule\end{tabular}\end{table}

\paragraph{Joint supervision balances semantic and quality learning.}
The full objective achieves 72.00\% gIoU and 77.01\% cIoU. Compared with removing whole-mask supervision, it gains 0.29\% and 0.43\%, respectively. Region and adoption supervision provide complementary improvements. Target-average and pooled-only losses favor their corresponding metrics, while their combination achieves a balanced result. Auxiliary weights between 0.05 and 0.2 produce closely grouped scores, supporting the chosen weighting.

\subsection{Candidate Mask and Fragmented Evidence Integration}
\label{sec:supp_integration}
Table~\ref{tab:consolidation} tests the effectiveness and advantage of our approach to integrating candidate masks and fragmented evidence over general aggregation strategies. Whole-Mask Selection uses the candidate as a unit, while union and majority voting follow overlap rules. Membership Regions support finer decisions conditioned on the complete instruction.
\begin{table}[!htbp]\centering
\caption{\textbf{Candidate Mask and Fragmented Evidence Integration.} Comparison with general aggregation strategies and ablations of our consolidation design. gIoU and cIoU evaluate the complete LaSeRS pipeline. BWG$_P$ evaluates the same variants on fixed high-potential ReVis-Eval banks.}\label{tab:consolidation}
\small\setlength{\tabcolsep}{4.5pt}
\begin{tabularx}{\linewidth}{L{1.9}*{3}{C{0.7}}}\toprule
\tablehead{Integration Strategy} & \tablehead{gIoU} & \tablehead{cIoU} & \tablehead{BWG$_P$}\\\midrule
Whole-Mask Selection & 69.19 & 73.43 & $-$1.46\\
Majority Vote & 68.41 & 72.80 & $-$0.93\\
Union & 66.84 & 71.93 & $-$3.37\\
Subset Union & 68.99 & 73.19 & +0.28\\
Independent Regions & 69.17 & 74.05 & +1.11\\
\midrule
Current Round Only & 71.44 & 76.32 & +1.72\\
Random Initialization & 70.63 & 75.10 & $-$0.08\\
No Adoption Selection & 71.01 & 75.32 & +2.27\\
\rowcolor{paperteal!9}\textbf{Full Consolidation} & 72.00 & 77.01 & +2.61\\
\bottomrule\end{tabularx}\end{table}

Independent Regions uses the same Shared Encoder, Membership Regions, and image and language context as the full model. It retains the same reference-mask initialization and Relative Adoption. Both variants use identical training samples, supervision labels, losses, and training steps.

The sole change is removing the Area-Weighted Summary $\bar z$ from the regional readout in Eq.~\ref{eq:supp_joint_readout}. Each region retains its contextualized feature $z_i$ but receives no pooled features from other regions.

Full consolidation exceeds Whole-Mask Selection by 2.81\% gIoU and 3.58\% cIoU. Joint composition also improves over Independent Regions by 2.83\% and 2.96\%. These gains support the connection in Eq.~\ref{eq:supp_netgain}: useful regional decisions should be judged by their combined effect on the complete mask.

Retaining all rounds adds 0.56\% gIoU over current-round evidence. Initializing from the learned selector adds 1.37\% over random initialization, and Relative Adoption adds 0.99\% over adopting every reconstruction. Together, these results support historical evidence retention, transfer of comparative understanding, and selection after consolidation.

\subsection{Revision Rounds and Computational Cost}
\label{sec:supp_cost}
\begin{figure*}[!t]
\centering\includegraphics[width=\textwidth]{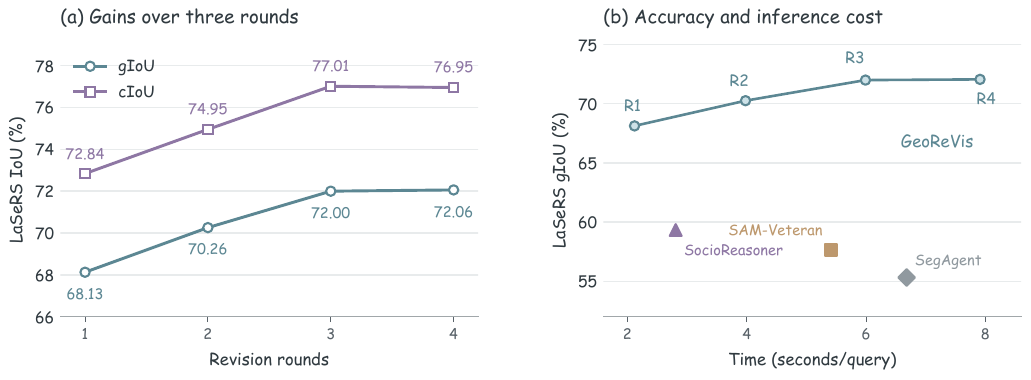}
\caption{\textbf{Accuracy and computational cost.} Left: LaSeRS gIoU and cIoU over one to four rounds. Right: gIoU versus amortized time on four A6000 GPUs. Baselines use their native backbones and three revision rounds. Two-round GeoReVis exceeds their accuracy while taking less time than SAM-Veteran and SegAgent. Three rounds improve accuracy further. Values come from Tables~\ref{tab:supp_rounds} and~\ref{tab:supp_efficiency} and main Table~1.}
\label{fig:supp_efficiency}
\vspace{12pt}
\nextfloat
\centering\includegraphics[width=\textwidth]{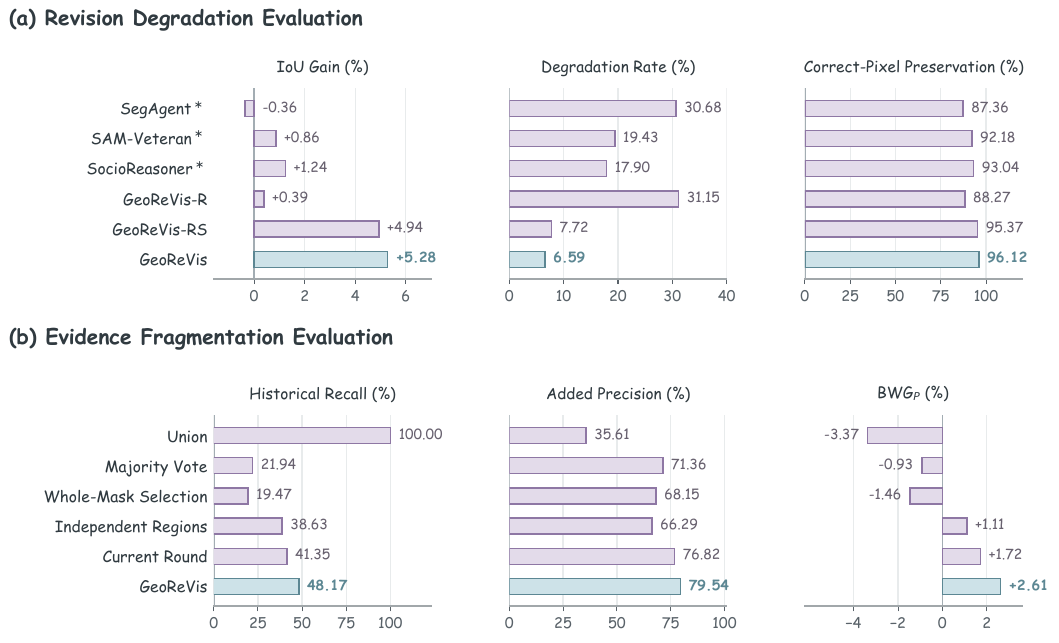}
\caption{\textbf{The two process capabilities measured by ReVis-Eval.} (a) Identical initial masks reveal how revision, selection, and consolidation affect final improvement and preservation. (b) Identical candidate banks reveal whether historical evidence is recovered accurately and whether the output improves beyond the optimal candidate. R and RS follow the main paper. All values are from Tables~\ref{tab:supp_full_revis} and~\ref{tab:supp_full_revis_b}.}
\label{fig:supp_process}
\end{figure*}

Figure~\ref{fig:supp_efficiency} and Tables~\ref{tab:supp_rounds}--\ref{tab:supp_efficiency} summarize accuracy and cost. One to three rounds improve gIoU from 68.13\% to 70.26\% and 72.00\%, with corresponding cIoUs of 72.84\%, 74.95\%, and 77.01\%. A fourth gives 72.06\% gIoU. Three rounds capture nearly all of the observed gIoU gain.
\begin{table}[!htbp]\centering
\caption{LaSeRS results (gIoU/cIoU) and amortized time (s/query) by revision rounds on four RTX A6000 GPUs. GeoReVis-RS uses revision and Candidate Selection, including the Concept Candidate.}\label{tab:supp_rounds}
\footnotesize\setlength{\tabcolsep}{3pt}
\resizebox{\linewidth}{!}{\begin{tabular}{lcccc}\toprule
\tablehead{Rounds} & \tablehead{1} & \tablehead{2} & \tablehead{3} & \tablehead{4}\\\midrule
GeoReVis-RS & 67.90/72.81 & 68.75/73.45 & 69.19/73.43 & 69.14/73.40\\
\rowcolor{paperteal!9}GeoReVis & 68.13/72.84 & 70.26/74.95 & 72.00/77.01 & 72.06/76.95\\
Time (s/query) & 2.12 & 3.98 & 5.99 & 7.91\\\bottomrule
\end{tabular}}\end{table}

\begin{table}[!htbp]\centering
\caption{Inference cost on LaSeRS with four RTX A6000 GPUs and three revision rounds. Time is amortized seconds per query. Calls and peak memory are per query and per GPU, respectively. Consolidation makes no additional segmenter calls.}\label{tab:supp_efficiency}
\footnotesize\setlength{\tabcolsep}{3pt}
\resizebox{\linewidth}{!}{\begin{tabular}{lccc}\toprule
\tablehead{Method} & \tablehead{Time (s/query)} & \tablehead{Calls} & \tablehead{Memory (GB)}\\\midrule
SocioReasoner$^\ddagger$ & 2.81 & 3.00 & 14.98\\
SAM-Veteran$^\ddagger$ & 5.41 & 4.07 & 24.68\\
SegAgent$^\ddagger$ & 6.68 & 7.64 & 23.23\\
GeoReVis w/o consolidation & 5.25 & 9.08 & 19.35\\
\rowcolor{paperteal!9}GeoReVis & 5.99 & 9.08 & 20.65\\
\bottomrule\end{tabular}}\end{table}

Two-round GeoReVis takes 3.98 seconds per query and exceeds all main-table baselines in gIoU. It is faster than SAM-Veteran at 5.41 seconds and SegAgent at 6.68 seconds. Three rounds add 1.74\% gIoU for a total of 5.99 seconds. Consolidation contributes 0.74 seconds per query.

\paragraph{Compact adapters specialize the execution roles.}
Each deployed Qwen3.5-2B instance contains 2.213B parameters, including its visual encoder. The seven rank-16 adapters contain 7.38M parameters each, totaling 51.67M. The regional readout adds 6.07M, including its frozen initialization branch. Our implementation retains the original and grounding-merged backbone instances, giving 4.484B resident VLM and readout parameters, excluding frozen SAM3. Roles reuse these instances through adapter switching, and each VLM call activates at most one adapter. Thus, role specialization preserves a compact per-call backbone compared with the 7B native backbones of SAM-Veteran and SegAgent~\cite{samveteran,segagent}.

\paragraph{Shared computation limits multi-stage overhead.}
Candidate Selection reads A/B preference logits directly. SAM3 caches image features across prompted decodes, while consolidation shares visual encoding across all Membership Regions and reconstructs the mask without additional segmenter calls. These mechanisms limit the cost of adding selection and evidence consolidation. Full GeoReVis uses 20.65\,GB peak memory per GPU, below SAM-Veteran's 24.68\,GB and SegAgent's 23.23\,GB (Table~\ref{tab:supp_efficiency}).

\subsection{Detailed ReVis-Eval Results}
\label{sec:supp_revis_details}
\paragraph{Selective revision improves feedback use and preservation.}
Table~\ref{tab:supp_full_revis} extends the main Revision Degradation Evaluation with Net Error Reduction. GeoReVis gains 5.28\% IoU and preserves 96.12\% of initially correct target pixels. Candidate Selection reduces Degradation Rate from 31.15\% for GeoReVis-R to 7.72\%, and full GeoReVis reaches 6.59\%. Figure~\ref{fig:supp_process}(a) makes this progression visible: improvement and preservation increase together.
\begin{table}[!htbp]\centering
\caption{\textbf{Revision Degradation Evaluation: complete results.} Shared initial masks isolate feedback revision. R denotes revision only. RS adds Candidate Selection, including the concept candidate. NER measures Net Error Reduction.}\label{tab:supp_full_revis}
\footnotesize\setlength{\tabcolsep}{2pt}
\resizebox{\linewidth}{!}{\begin{tabular}{lccccc}\toprule
\tablehead{Method} & \tablehead{$\Delta$IoU$\uparrow$} & \tablehead{FG$\uparrow$} & \tablehead{DR$\downarrow$} & \tablehead{CP$\uparrow$} & \tablehead{NER$\uparrow$}\\\midrule
SegAgent$^\ast$ & $-$0.36 & +0.43 & 30.68 & 87.36 & $-$1.62\\
SAM-Veteran$^\ast$ & +0.86 & +0.71 & 19.43 & 92.18 & 4.27\\
SocioReasoner$^\ast$ & +1.24 & +0.86 & 17.90 & 93.04 & 5.83\\
GeoReVis-R & +0.39 & +1.12 & 31.15 & 88.27 & 2.84\\
GeoReVis-RS & \underline{+4.94} & \underline{+1.95} & \underline{7.72} & \underline{95.37} & \underline{17.86}\\
\rowcolor{paperteal!9}\textbf{GeoReVis} & \textbf{+5.28} & \textbf{+2.08} & \textbf{6.59} & \textbf{96.12} & \textbf{19.43}\\
\bottomrule\end{tabular}}
\end{table}

\begin{table}[!htbp]\centering
\caption{\textbf{Evidence Fragmentation Evaluation: complete results.} Every method receives the same candidate banks. BWG compares the output with the optimal candidate. BWG$_P$ uses high-potential banks. HR and AP measure historical recovery and Added Precision.}\label{tab:supp_full_revis_b}
\footnotesize\setlength{\tabcolsep}{2pt}
\resizebox{\linewidth}{!}{\begin{tabular}{lccccc}\toprule
\tablehead{Method} & \tablehead{IoU$\uparrow$} & \tablehead{BWG$\uparrow$} & \tablehead{BWG$_P\uparrow$} & \tablehead{HR$\uparrow$} & \tablehead{AP$\uparrow$}\\\midrule
Keep Reference & 66.50 & $-$2.87 & $-$6.63 & 0.00 & --\\
SAM Score & 66.17 & $-$3.20 & $-$6.09 & 8.71 & 57.42\\
Prompted Qwen & 64.69 & $-$4.68 & $-$6.37 & 15.86 & 49.08\\
Union & 60.26 & $-$9.11 & $-$3.37 & 100.00 & 35.61\\
Majority Vote & 67.65 & $-$1.72 & $-$0.93 & 21.94 & 71.36\\
Whole-Mask Selection & 68.74 & $-$0.63 & $-$1.46 & 19.47 & 68.15\\
Subset Union & 69.00 & $-$0.37 & +0.28 & 31.82 & 62.74\\
Independent Regions & 68.66 & $-$0.71 & +1.11 & 38.63 & 66.29\\
Current Round & \underline{69.60} & \underline{+0.23} & \underline{+1.72} & \underline{41.35} & \underline{76.82}\\
\rowcolor{paperteal!9}\textbf{GeoReVis} & \textbf{70.41} & \textbf{+1.04} & \textbf{+2.61} & \textbf{48.17} & \textbf{79.54}\\
\midrule
Optimal Candidate & 69.37 & 0.00 & 0.00 & -- & --\\
Oracle Subset Union & 70.73 & +1.36 & +2.69 & -- & --\\
Regional Oracle & 75.76 & +6.39 & +15.12 & -- & --\\
\bottomrule\end{tabular}}
\end{table}

\paragraph{Consolidation recovers precise evidence beyond the optimal candidate.}
The complete Evidence Fragmentation Evaluation accompanies the revision results. Full GeoReVis reaches +1.04\% BWG overall and +2.61\% on high-potential banks. Figure~\ref{fig:supp_process}(b) shows that its Historical Recall of 48.17\% is accompanied by 79.54\% Added Precision. Compared with current-round evidence, full history adds 6.82\% Historical Recall and 2.72\% Added Precision. The additional evidence is both recovered and more accurately selected.

Table~\ref{tab:supp_revis_a} separates the three initial-mask generators. GeoReVis improves all three starting distributions, with gains of 3.88\%, 6.81\%, and 5.02\%. Their sample-weighted average is the 5.28\% gain reported in the main paper. This shows that the feedback and selection mechanism works across distinct sources of initial errors.

Table~\ref{tab:supp_revis_b} separates high-potential banks from the remaining banks. GeoReVis achieves +2.61\% BWG on high-potential banks and +0.56\% on the remaining banks, yielding +1.04\% overall. Full GeoReVis also preserves 98.06\% of correct reference pixels. The controlled banks link the gain to improved evidence use rather than additional segmentation calls.
\begin{table}[!htbp]\centering
\caption{ReVis-Eval Revision Degradation Evaluation by initial-mask generator: $\Delta$IoU / DR. Items (mean initial IoU): GeoReVis 1,021 (61.29), VisionReasoner$^\ddagger$ 1,108 (50.46), SocioReasoner$^\ddagger$ 1,072 (54.03).}\label{tab:supp_revis_a}
\footnotesize\setlength{\tabcolsep}{2pt}
\resizebox{\linewidth}{!}{\begin{tabular}{lccc}\toprule
\tablehead{Method} & \tablehead{GeoReVis} & \tablehead{VisionReasoner$^\ddagger$} & \tablehead{SocioReasoner$^\ddagger$}\\\midrule
SegAgent$^\ast$ & $-$0.83 / 33.50 & +0.12 / 27.62 & $-$0.41 / 31.16\\
SAM-Veteran$^\ast$ & +0.41 / 21.84 & +1.47 / 16.88 & +0.66 / 19.78\\
SocioReasoner$^\ast$ & +0.74 / 19.29 & +1.86 / 15.88 & +1.09 / 18.66\\
GeoReVis-R & $-$0.12 / 33.99 & +1.03 / 28.43 & +0.21 / 31.25\\
GeoReVis-RS & +3.62 / 8.91 & +6.37 / 6.50 & +4.71 / 7.84\\
\rowcolor{paperteal!9}\textbf{GeoReVis} & +3.88 / 7.64 & +6.81 / 5.32 & +5.02 / 6.90\\
\bottomrule\end{tabular}}\end{table}

\begin{table}[!htbp]\centering
\caption{ReVis-Eval Evidence Fragmentation Evaluation by stratum. High-potential banks (812/3,474): regional oracle exceeds the best complete mask by $>5\%$. $\mathrm{CP}_E$: Correct-Pixel Preservation relative to the reference.}\label{tab:supp_revis_b}
\footnotesize\setlength{\tabcolsep}{2.4pt}
\begin{tabular}{lcccc}\toprule
\tablehead{Method} & \tablehead{BWG$_P$} & \tablehead{BWG$_{\bar P}$} & \tablehead{BWG} & \tablehead{CP$_E$}\\\midrule
Keep Reference & $-$6.63 & $-$1.72 & $-$2.87 & 100.00\\
SAM Score & $-$6.09 & $-$2.32 & $-$3.20 & 92.63\\
Prompted Qwen & $-$6.37 & $-$4.16 & $-$4.68 & 89.27\\
Union & $-$3.37 & $-$10.86 & $-$9.11 & 100.00\\
Majority Vote & $-$0.93 & $-$1.96 & $-$1.72 & 97.62\\
Whole-Mask Selection & $-$1.46 & $-$0.38 & $-$0.63 & 95.84\\
Subset Union & +0.28 & $-$0.57 & $-$0.37 & 96.47\\
Independent Regions & +1.11 & $-$1.27 & $-$0.71 & 95.91\\
Current Round & +1.72 & $-$0.22 & +0.23 & 97.48\\
\rowcolor{paperteal!9}\textbf{GeoReVis} & +2.61 & +0.56 & +1.04 & 98.06\\
Oracle Subset Union & +2.69 & +0.95 & +1.36 & --\\
Regional Oracle & +15.12 & +3.73 & +6.39 & --\\
\bottomrule\end{tabular}\end{table}

\subsection{Cross-Mask Evidence Recovery Case Study}
\label{sec:supp_recovery_case}
\begin{figure*}[!t]
\centering\includegraphics[width=\textwidth]{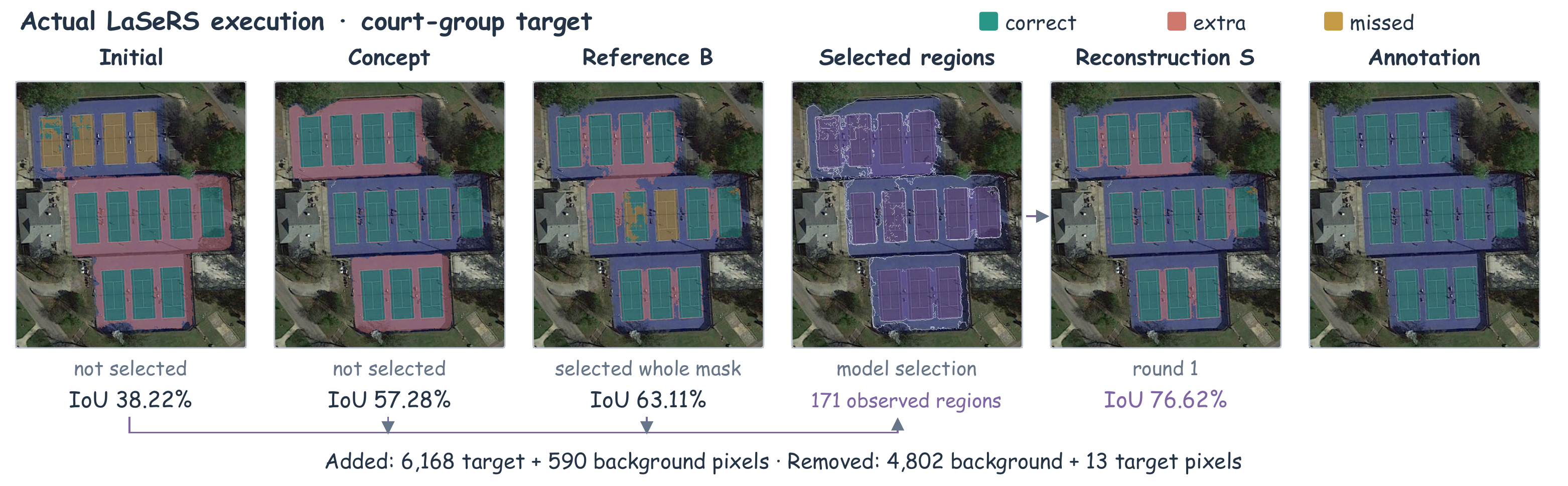}
\caption{\textbf{Actual recovery of fragmented evidence.} An intermediate GeoReVis checkpoint reconstructs a 76.62\%-IoU mask from candidates whose best IoU is 63.11\% on LaSeRS image 21111. Relative to the reference, it adds 6,168 target and 590 background pixels, and removes 4,802 background and 13 target pixels.}
\label{fig:actual}
\vspace{6pt}
\nextfloat
\centering\includegraphics[width=0.60\textwidth]{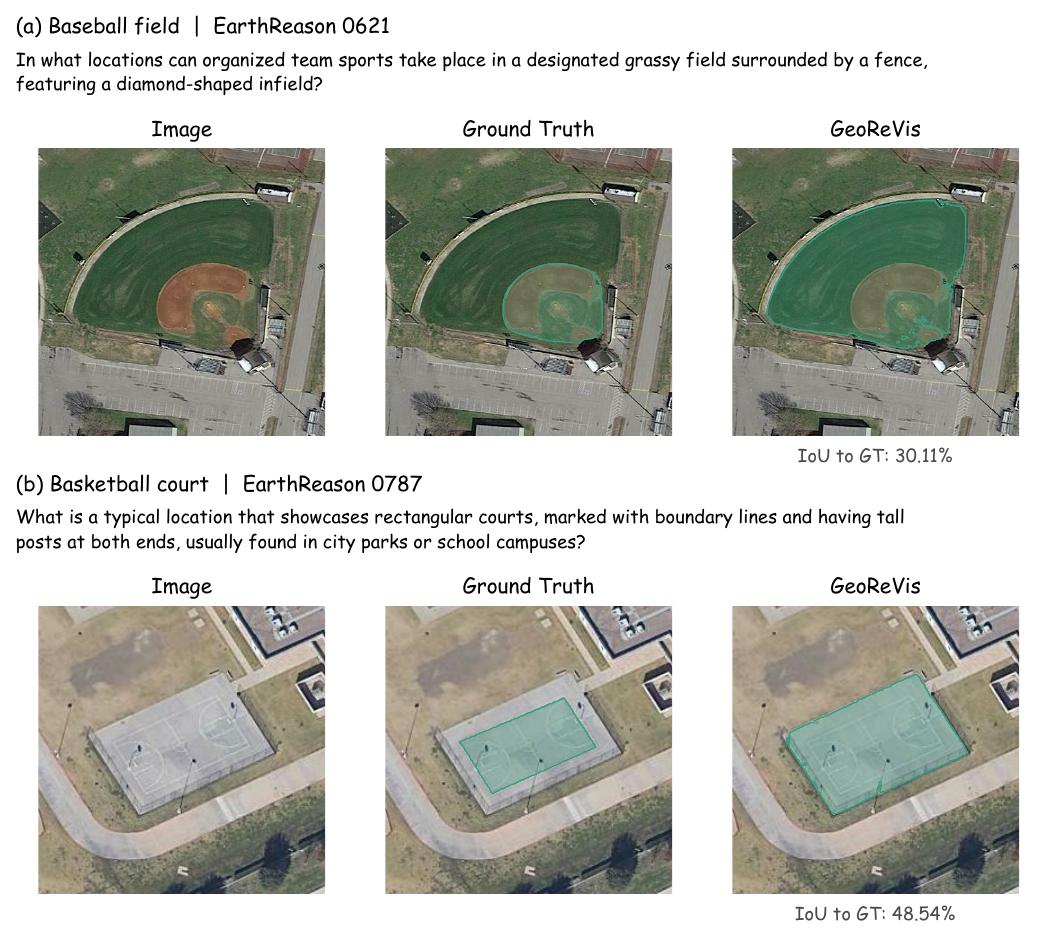}
\caption{\textbf{Target coverage beyond the supplied annotation on EarthReason.} Original images, unchanged GT masks, and actual predictions are shown with the test instructions. Green marks foreground. In (a), the prediction includes the described grassy outfield, while GT mainly covers the infield. In (b), the prediction covers more of the visible basketball court. Displayed IoUs measure agreement with the original annotations.}
\label{fig:earth_annotation}
\end{figure*}

Figure~\ref{fig:actual} shows an actual execution on LaSeRS image 21111. The requested court group is partly covered by several candidates. A Concept Candidate contains useful courts together with substantial background, so whole-mask selection rejects it. Joint composition recovers the useful regions while excluding much of that background, reaching 76.62\% IoU compared with 63.11\% for the optimal source candidate.
\begin{table*}[!t]
\centering
\caption{\textbf{Influence of backbone and segmenter on GeoReVis.} Scores are gIoU (\%). EarthReason averages Val and Test gIoU. LaSeRS uses its official nine-dimension macro gIoU, and referring benchmarks use test gIoU. Avg. is the equally weighted mean across the five benchmarks. SOTA scores follow main Tables~1--3 and Supplementary Tables~\ref{tab:supp_earth}--\ref{tab:supp_referring}. Best and second-best scores are bold and underlined within each block.}
\label{tab:supp_backbone_segmenter}
\footnotesize\setlength{\tabcolsep}{2.5pt}
\begin{tabularx}{\textwidth}{L{1.4}L{1.75}C{1.05}C{0.70}C{0.95}C{0.80}C{0.80}C{0.85}C{0.70}}
\toprule
\tablehead{Method} & \tablehead{Backbone} & \tablehead{Segmenter} & \tablehead{LaSeRS} & \tablehead{EarthReason} & \tablehead{RRSIS-D} & \tablehead{RefSegRS} & \tablehead{RISBench} & \tablehead{Avg.}\\
\midrule
\multicolumn{9}{c}{\textbf{\textit{SOTA Methods}}}\\
\midrule
GeoReason~\cite{georeason} & Qwen3-VL-4B & SAM2 & \textbf{64.96} & \textbf{78.11} & 67.32 & 76.38 & \underline{71.36} & \textbf{71.63}\\
UniGeoSeg~\cite{unigeoseg} & Phi-1.5 + Swin-B & Mask2Former & & \underline{72.81} & \textbf{69.25} & 76.06 & 70.94 & \\
CROSS~\cite{cross} & SigLIP 2 & SAM2 & -- & -- & 68.92 & \textbf{79.51} & 70.86 & --\\
RSRefSeg 2~\cite{rsrefseg2} & SigLIP 2 & SAM2 & -- & -- & \underline{69.17} & \underline{77.39} & \textbf{72.57} & --\\
\midrule
GeoReVis & Qwen2.5-VL-7B & SAM2 & 70.94 & 78.20 & 69.51 & 79.82 & 72.63 & 74.22\\
GeoReVis & Qwen3.5-2B & SAM2 & 71.07 & 78.22 & 70.01 & \underline{80.49} & 73.26 & 74.61\\
GeoReVis & Qwen2.5-VL-7B & SAM3 & \underline{71.36} & \underline{78.35} & \textbf{70.42} & 79.88 & \underline{73.98} & \underline{74.80}\\
\rowcolor{paperteal!9}GeoReVis & Qwen3.5-2B & SAM3 & \textbf{72.00} & \textbf{78.46} & \underline{70.35} & \textbf{80.72} & \textbf{74.27} & \textbf{75.16}\\
\bottomrule
\end{tabularx}
\end{table*}

The pixel changes explain the gain: 6,168 missing target pixels are recovered, while 4,802 background pixels are removed. This example demonstrates the intended mechanism at the region level. A candidate's rejection as a complete answer does not discard the useful evidence it contains.

\subsection{Qualitative Analysis on EarthReason}
\label{sec:supp_earth_qual}
Figure~\ref{fig:earth_annotation} shows two three-round outputs from an intermediate LaSeRS-trained checkpoint. Table~\ref{tab:supp_earth} reports benchmark-trained scores.

\paragraph{Complete instructions support more complete target coverage.}
In sample 0621, GeoReVis covers the grassy outfield surrounding the annotated infield. In sample 0787, it covers more of the described court. The predictions retain 98.85\% and 100.00\% of annotated foreground, respectively.

Their annotation IoUs are 30.11\% and 48.54\%, respectively.

\paragraph{Annotation quality affects execution-driven supervision.}
Some EarthReason annotations have coarse target boundaries or incomplete target coverage. During benchmark-specific training, the Execution-Driven Data Pipeline derives action utility, candidate preferences, and region composition targets from these annotations.

Correct target pixels outside an annotated boundary can be penalized, while incorrectly annotated background can be rewarded. Such errors introduce noise into the supervision for revision, selection, and consolidation. The examples above nevertheless show that GeoReVis can delineate the requested targets more faithfully than the supplied GT masks.

\subsection{Influence of Backbone and Segmenter}
\label{sec:supp_backbone_segmenter}
Table~\ref{tab:supp_backbone_segmenter} compares four backbone and segmenter combinations under the same three-round GeoReVis framework and benchmark-specific training protocol. All four outperform the strongest SOTA baseline on each benchmark, while their average gIoU varies by just 0.94\%. These results support the consistent effectiveness of Selective Feedback Revision, Cross-Mask Evidence Consolidation, and the Execution-Driven Data Pipeline across different backbones and segmenters.

We adopt Qwen3.5-2B with SAM3 as the default configuration. It achieves the highest average gIoU of 75.16\% with a smaller language backbone than the 7B alternatives.

\newpage
\begin{samepage}
\section{Limitations}
\label{sec:supp_limitations}
\paragraph{Unobserved target regions.}
The Execution-Driven Data Pipeline provides reliable supervision, Selective Feedback Revision acquires useful target evidence, and Cross-Mask Evidence Consolidation combines complementary regions. However, consolidation is restricted to the union of observed candidates (Sec.~\ref{sec:supp_representation}). If a target region is absent from every candidate throughout Initial Segmentation and all revision rounds, GeoReVis cannot recover it through consolidation.

\paragraph{Computational cost.}
Two-round GeoReVis outperforms all baselines in main Table~1 and runs faster than SAM-Veteran and SegAgent. The default three-round model improves accuracy further but increases inference time from 3.98 to 5.99 seconds per query (Tables~\ref{tab:supp_rounds}--\ref{tab:supp_efficiency}). Its cost exceeds SocioReasoner and SAM-Veteran, leaving room to improve the accuracy--efficiency trade-off.

\paragraph{Saturation across revision rounds.}
GeoReVis improves segmentation through feedback revision and accumulated evidence, but gains become marginal beyond three rounds (Table~\ref{tab:supp_rounds}). Annotation errors affect the quality of execution-driven supervision, while the model's ability to discover and combine target evidence remains bounded. Repeating inference alone therefore does not provide unlimited improvement. We view Recursive Self-Improvement (RSI) as a potential direction for extending this capability beyond the current model's limits.
\par
\end{samepage}

\end{document}